\documentclass[journal]{IEEEtran}
\usepackage{amsmath,amsfonts}
\usepackage{algorithmic}
\usepackage{algorithm}
\usepackage{array}
\usepackage{bm}
\usepackage{booktabs}
\usepackage{textcomp}
\usepackage{stfloats}
\usepackage{multirow}
\usepackage{url}
\usepackage{verbatim}
\usepackage{cite}
\usepackage{graphicx}
\usepackage{subfigure}
\usepackage{threeparttable}
\usepackage{lineno}
\usepackage{enumitem}
\usepackage{soul,color,xcolor}
\usepackage{hyperref}
\usepackage{float}
\begin{document}

\title{\LARGE \bf Learning-Based Leader Localization for Underwater Vehicles With Optical-Acoustic-Pressure Sensor Fusion}

\author{Mingyang Yang, Zeyu Sha and Feitian Zhang$^*$
\thanks{The authors are with the Department of Advanced Manufacturing and Robotics, College of Engineering, and the State Key Laboratory of Turbulence and Complex Systems, Peking University, Beijing, 100871, China (emails:
        {mingyangyang@stu.pku.edu.cn},  {schahzy@stu.pku.edu.cn} and {feitian@pku.edu.cn}).}%
\thanks{* Send all correspondence to Feitian Zhang.}
\thanks{The dataset is available at: https://github.com/MrRyanYang/Tri-Fusion.}
}

\maketitle

\begin{abstract}
Underwater vehicles have emerged as a critical technology for exploring aquatic environments. The deployment of multi-vehicle systems has gained substantial interest due to their capability to perform collaborative tasks with improved efficiency. However, achieving precise localization of a leader underwater vehicle within a multi-vehicle configuration remains a significant challenge, particularly in dynamic and complex underwater conditions. 
To address this issue, this paper presents a novel tri-modal sensor fusion neural network approach that integrates optical, acoustic, and pressure sensors to localize the leader vehicle. The proposed method leverages the unique strengths of each sensor modality to improve localization accuracy and robustness. Specifically, optical sensors provide high-resolution imaging for precise relative positioning, acoustic sensors enable long-range detection and ranging, and pressure sensors offer environmental context awareness. The fusion of these sensor modalities is implemented using a deep learning architecture designed to extract and combine complementary features from raw sensor data. The effectiveness of the proposed method is validated through a custom-designed testing platform and field test. Extensive data collection and experimental evaluations demonstrate that the tri-modal approach significantly improves the accuracy and robustness of leader localization, outperforming both single-modal and dual-modal methods.
\end{abstract}


\section{Introduction}
\IEEEPARstart{U}{nderwater} vehicles, including Autonomous Underwater Vehicles (AUVs) and Remotely Operated Vehicles (ROVs), have revolutionized the exploration of aquatic environments
\cite{palomeras2019autonomous}, \cite{verfuss2019review},
\cite{sward2019systematic}, \cite{2022RAL-monitor}. These vehicles provide unprecedented access to deep-sea regions that are otherwise inaccessible to human divers or traditional survey methods. As the capabilities of these vehicles continue to evolve, multi-vehicle systems, also known as underwater vehicle swarms, have emerged as a promising approach to enhance the efficiency of underwater missions. Multi-vehicle systems, comprising multiple underwater vehicles operating collaboratively, enable a wide range of tasks, such as seabed mapping \cite{Galceran2012seabed}, \cite{wang2023autonomous}, 
environmental monitoring 
\cite{lodovisi2018performance}, 
and marine exploration 
\cite{zhang2021auv}, 
\cite{li2024method}. By leveraging the strengths of individual underwater vehicles, swarms typically cover larger areas, share computational resources, and improve mission resilience 
\cite{wang2022task}, 
\cite{chen2020multi}.

Despite these advantages, a critical challenge in deploying underwater vehicle swarms is achieving effective relative localization of leader vehicles within the multi-agent system. Accurate relative localization is crucial for maintaining formation stability, ensuring safe navigation, and enabling precise coordination among vehicles \cite{li2019localization}, \cite{wei2023localization}. 
Researchers have investigated various techniques for relative localization, including optical, acoustic, and pressure sensor-based approaches 
\cite{cong2021underwater}, 
\cite{huy2023object}. Optical sensors, such as cameras, provide high-resolution imaging for precise relative positioning in close proximity \cite{2024HongVision}, \cite{2019ManzanVision}. Acoustic sensors, on the other hand, offer long-range detection and are well-suited for environments with limited visibility 
\cite{2020MachadoAcoustic}, 
\cite{xu2023localization}. Pressure sensors provide valuable data on depth and environmental context 
\cite{2024WangPressure}, \cite{ZhenglateralTRO}.

While each sensor modality provides distinct advantages, underwater environments are inherently dynamic and complex, characterized by time-varying currents, limited visibility, and obstacles, which limit the performance and robustness of single-sensor systems. Researchers have explored dual-modality methods to address these limitations by combining two sensor types to improve localization accuracy \cite{Hu2021VisionPressure,2024MaVisionPressure}. In recent years, with the development of development of deep learning methods, a growing community of researchers have applied learning-based and data-driven methods to underwater sensor data fusion field in order to realize accurate target positioning. The underwater dual-modality sensor fusion is categorized into mainly three types \cite{huy_object_2023}.
The first type of fusion involves fusion using acoustic sensors only. Wang \textit{et al.} \cite{wang2023underwater} proposed a feature fusion framework to combine the intensity features extracted from the polar image representation and the geometric features learned from the point cloud representation of sonar data. Sadjoli \textit{et al.} \cite{sadjoli2023pcd} proposed novel perception-based applications for Orthogonal Multibeam Sonar Fusion (OMSF), consisting of classification technique integrating Point Cloud Data (PCD) with a PCD-based Convolutional Neural Network (CNN), and pose estimation method combining orthogonal feature matching bounding box regression with a pose regression CNN.
The second type of fusion emerges between acoustic sensor and another type of sensor \cite{Song2024VisionAcoustic}. Yang \textit{et al.} \cite{Yang2024VisionAcoustic} proposed a localization method assimilating optical and acoustic measurements, while Dos Santos \textit{et al.} \cite{2020DosSantosMatching} proposed a cross-domain and cross-view image matching, using aerial images and underwater acoustic images to assist underwater localization.
The third type of fusion occurs without acoustic sensors. Jiang \textit{et al.} \cite{jiang2021underwater} developed an artificial lateral line system integrated with pressure sensors and flow velocity sensors to realize source localization based on multilayer perceptron neural network. Bongiorno \textit{et al.} \cite{bongiorno2018coregistered} developed a new method for high-resolution hyperspectral mapping of the seafloor utilizing a spectrometer co-registered with a high-resolution color stereo camera system onboard an AUV.

However, dual-modality systems, while advantageous over single-modality, often fail to fully exploit the complementary nature of all available sensor types. This limitation underscores the potential of tri-modal fusion methods. By integrating data from optical, acoustic, and pressure sensors, these methods offer a more holistic and accurate perception of the leader's position, advancing the state of the art in underwater relative localization. The tri-modal fusion method could be potentially applied other underwater robot swarm researches. In the research on long-term marine environment monitoring with a heterogeneous robotic swarm \cite{lonvcar2019heterogeneous}, the method could optimize the localization of underwater sensor nodes like aMussels, enhancing data collection accuracy.
In the case of the Blueswarm platform demonstrating 3D collective behaviors \cite{berlinger2021implicit}, the method could assist Bluebots in more accurate target localization during search operations.

\begin{figure}[thpb]
 \centering
 \includegraphics[width=0.48\textwidth]{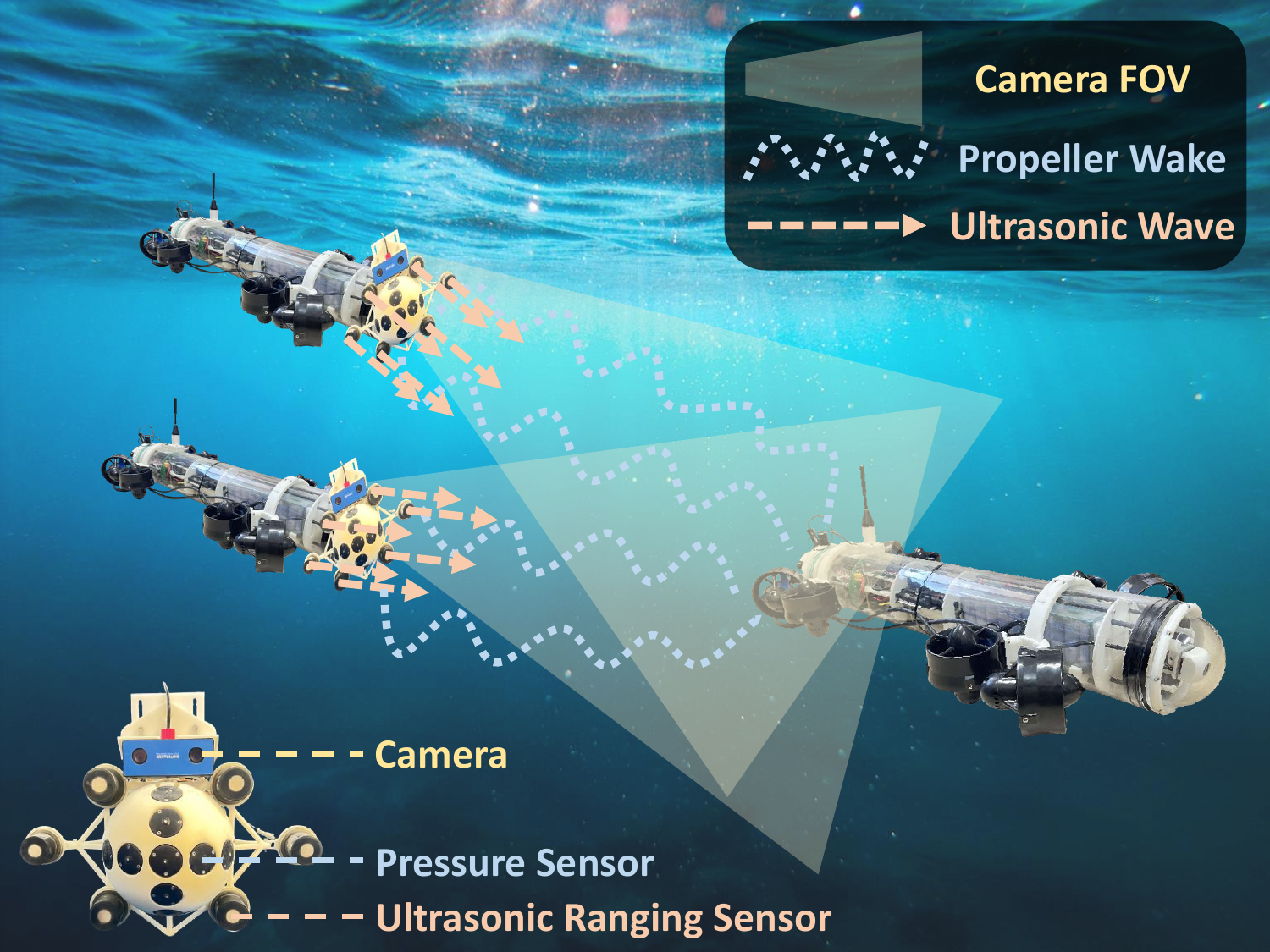}
 \caption{Illustration of an underwater vehicle swarm utilizing the proposed multi-modal leader vehicle localization framework. The visual camera operates within the camera's field of view (FOV), the acoustic ranging sensors function within their transmission and reception angles, and the pressure sensors respond to detectable propeller wake. Visual cameras (yellow), acoustic ranging sensors (orange), and pressure sensors (blue) collectively gather information about the leader vehicle to estimate its states of interest.}
\label{IntroFig}
\end{figure}

This paper presents a novel tri-modal fusion algorithm that integrates optical, acoustic, and pressure sensors to localize the leader underwater vehicle within a multi-vehicle system. An envisioned underwater vehicle swarm is illustrated in Fig.~\ref{IntroFig}. The proposed method leverages the complementary strengths of these sensor modalities to enhance the accuracy and robustness of leader localization. By fusing data from diverse sensors, the algorithm mitigates the limitations of individual modalities, providing reliable position estimates in challenging underwater environments.

The multi-modal sensor fusion is achieved through a deep learning architecture designed to extract and combine complementary features from  raw sensor data. Deep learning has demonstrated remarkable capabilities across various domains owing to its ability to learn complex patterns and relationships from large datasets \cite{xu2023systematic,fayyad2020deep,kuutti2020survey}. In this paper, deep learning enables effective fusion of multi-modal sensor data, extracting high-level features critical for accurate position estimation \cite{maurelli2022auv}, \cite{ding2024robust}. To validate the proposed method, this paper develops a comprehensive testing platform comprising a test pool, a tri-modal sensing module, and a target module. Extensive data collection and experimental evaluations are conducted to assess the algorithm's performance. The experimental results demonstrate that the tri-modal fusion deep learning method significantly improves position estimation performance and robustness compared to single- and dual-modal approaches.

The main contributions of this work are threefold. First, we propose a novel tri-modal fusion framework that integrates optical, acoustic, and pressure sensors for underwater leader localization, addressing the limitations of conventional single- and dual-modal approaches in dynamic environments. Unlike prior works, our method uniquely leverages the complementary strengths of all three modalities, i.e. optical imaging for precision, acoustic sensing for long-range robustness, and pressure measurements for flow information. Second, we design a lightweight end-to-end deep learning architecture that innovatively fuses modalities at both data and feature levels. Specifically, acoustic heatmaps are embedded into RGB images as an attention channel, while spatiotemporal pressure features are extracted via a hybrid neural network and fused with optical-acoustic features. This architecture is tailored to handle heterogeneous sensor data and their distinct physical characteristics. Third, we develop an underwater testing platform and conduct extensive experiments to validate our approach. The results demonstrate significant improvements in accuracy and robustness. These advancements address real-world challenges in underwater swarm coordination, offering a possible solution for missions in turbid or noisy underwater environments.

\section{Problem Description}
This section presents the problem addressed in this study and describes the development of the testing platform used for experimentally validating the proposed algorithm. 

\subsection{Testing Platform}
This paper develops and constructs a testing platform to faciliate the leader underwater vehicle localization experiments. The platform consists of three primary components, including a test pool, a target module and a tri-modal sensing module. The test pool measures 4~m, 2~m and 0.7~m in length, width and depth, respectively.
\begin{figure}[thpb]
\centering
\label{PhysicalEnv}
\subfigure[Target module.]{
\label{target}
\includegraphics[width=0.23\textwidth, angle=0]{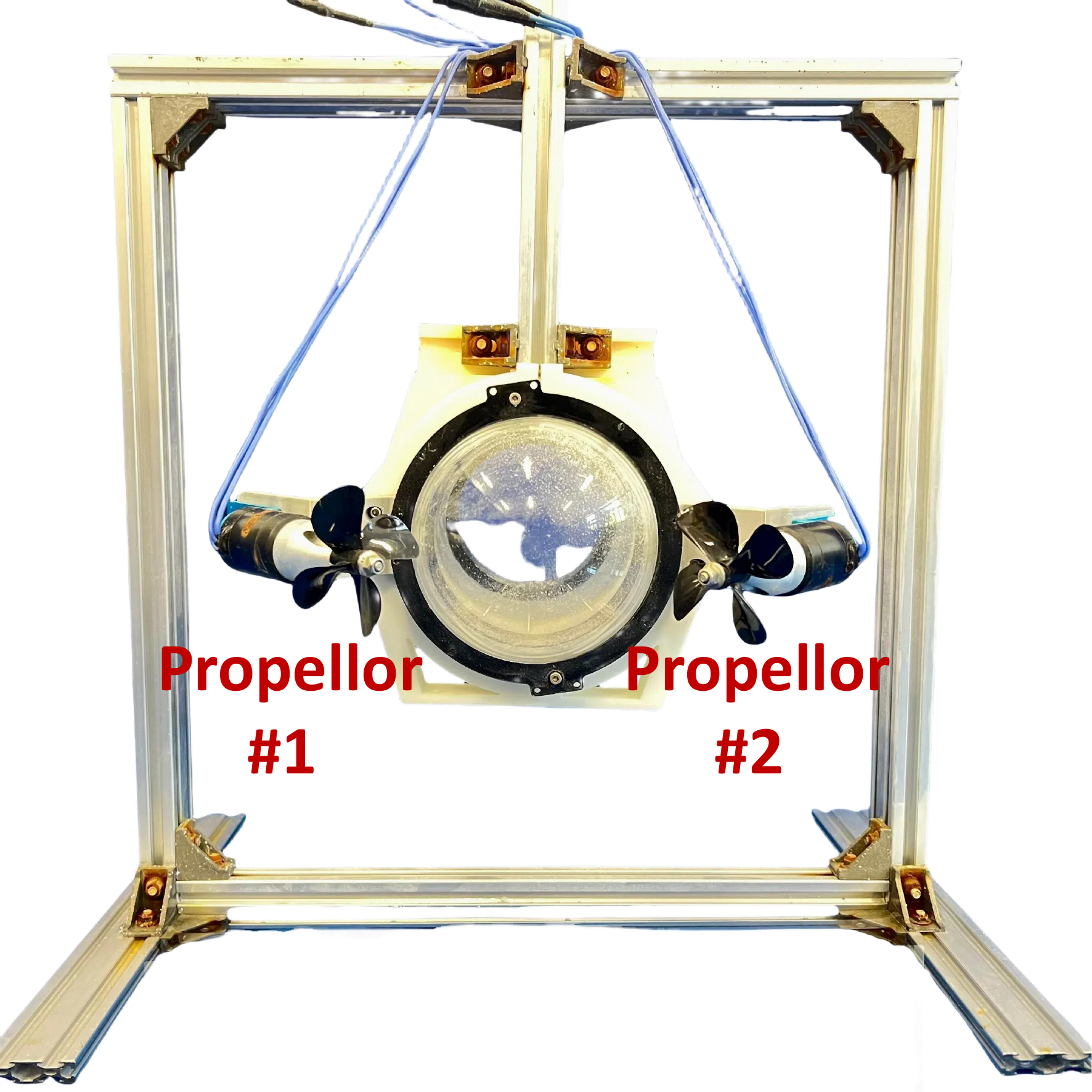}}
\subfigure[Sensing module.]{
\label{sensingmodule}
\includegraphics[width=0.23\textwidth]{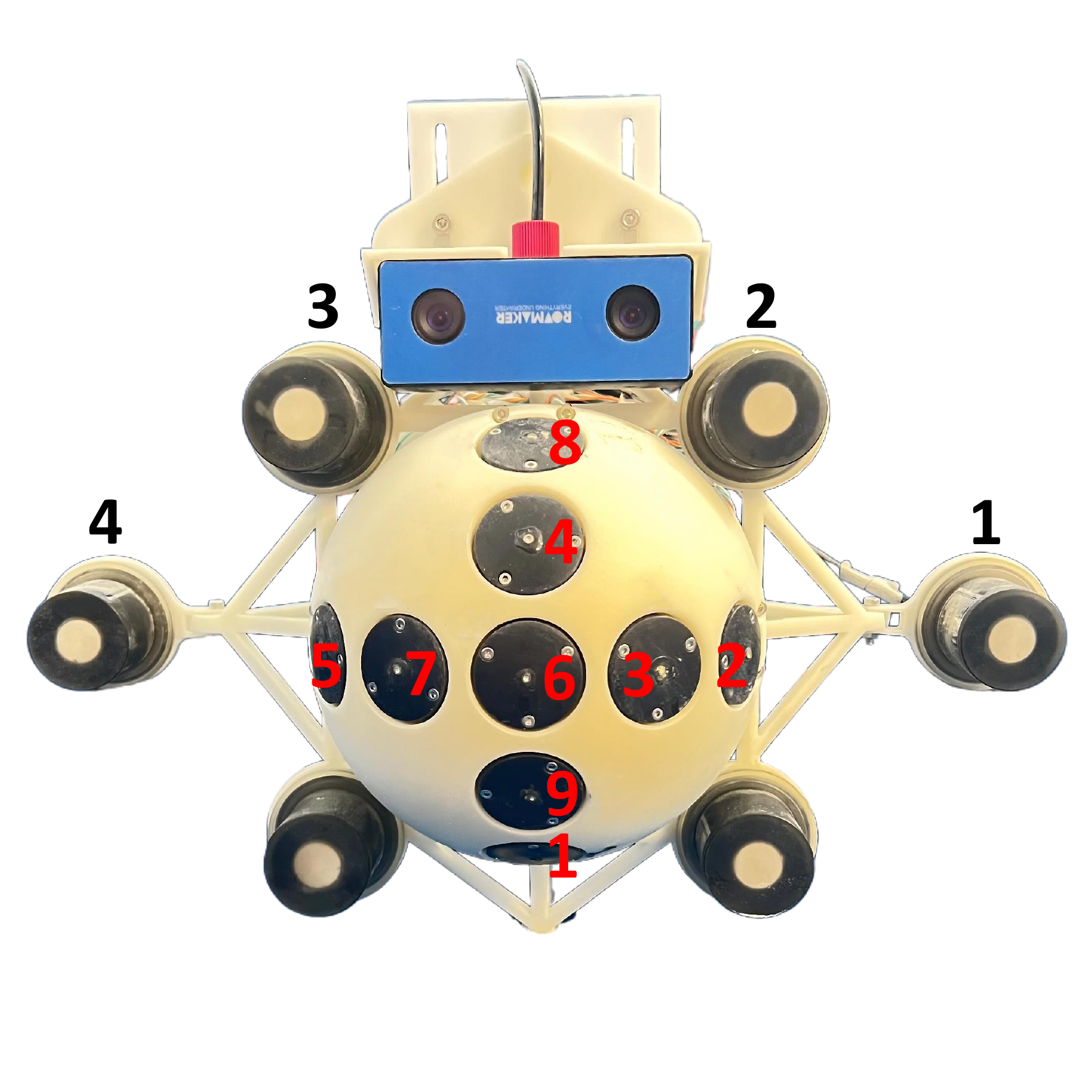}}
\caption{Illustration of the target module and the tri-modal sensing module. The target module (Fig.~\ref{target}) represents the tail section of the leader vehicle. The tri-modal sensing module (Fig.~\ref{sensingmodule}), resembling the head section of the follower vehicle, comprises a 3D-printed support structure, a binocular camera, nine pressure sensors (indexed in red), and six acoustic ranging sensors (indexed in black). Under proper actuation, the propellers in the target module rotate at a constant speed, simulating real-world conditions during underwater vehicle movement and turning. All the camera, acoustic ranging sensors, and the pressure sensors in the sensing module capture the associate data, which is subsequently collected for target module localization.}
\end{figure}

The target module, representing the tail section of the leader vehicle and modeled after the OpenAUV, a lab-developed underwater robot \cite{OpenAUV}, is supported by an aluminum profile frame, as illustrated in Fig.~\ref{target}. To capture propeller wake data, two waterproof underwater thrusters (Pasotim 2838) are employed as propellers. The propellers are powered via two Skywolf TL-80 servo drivers, which regulate their rotation direction and speed through pulse-width modulation. The pulse width ranges from 1K to 2K \textmu s, corresponding to the lower and upper control limits, respectively.

The multi-modal sensing module, resembling the head section of the follower vehicle, is secured using an acrylic holder, as illustrated in Fig.~\ref{sensingmodule}. This module comprises a 3D-printed support structure, a binocular camera mounted at the top, nine pressure sensors centrally located, and six acoustic ranging sensors arranged peripherally. The binocular camera, manufactured by ROVMAKER, supports various resolutions, with a maximum resolution of 2560$\times$960 pixels. The acoustic ranging sensors are 300~KHz Yuzheng04 modules, manufactured by Hangzhou Umbrella Automation Technology. The pressure sensors used are MS5837-02BA, featuring a resolution of 1~Pa and communication via the I2C bus. Real-time acquisition of data from the nine pressure sensors is achieved using three I2C multiplexers from DFRobot. To ensure waterproofing, the MS5837-02BA sensors are enclosed within protective structures. A Raspberry Pi 4B with 8~GB memory is utilized for data acquisition, processing and storage.

\subsection{Localization Problem}
In this study, the leader and follower vehicles are assumed to operate at the same depth, restricting the leader localization problem to the horizontal plane. The test pool features a water depth of 50~cm, with the target module and the multi-modal sensing module positioned 25~cm below the water surface.
\begin{figure}[thpb]
 \centering
 \includegraphics[width=0.49\textwidth]{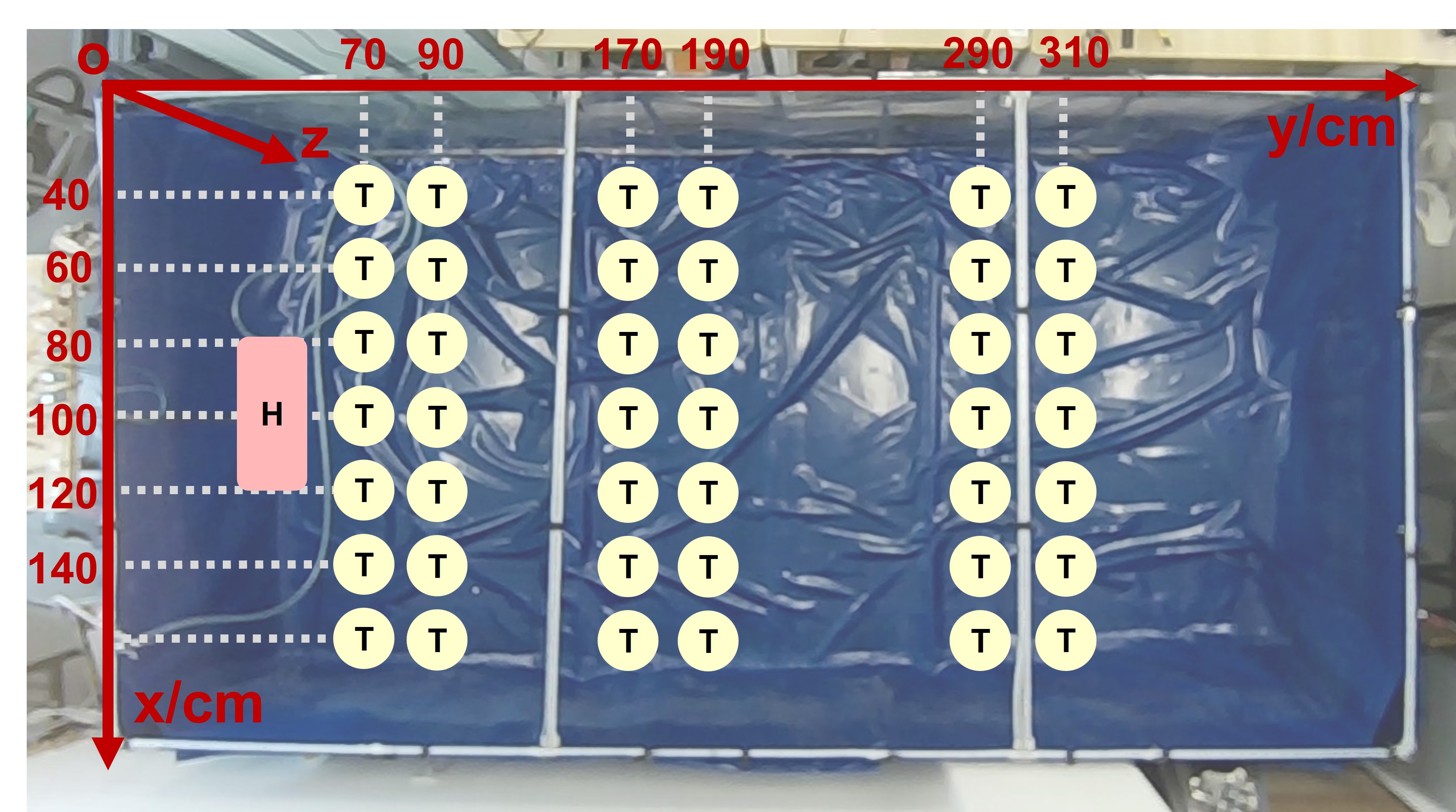}
 \caption{Illustration of the sample locations within the test pool (top view). A mesh grid comprising 42 locations is selected. The fixed multi-modal sensing module is denoted as \textbf{H}, while the target module, represented as \textbf{T}, is manually positioned at each grid vertex. For each location, three orientations of the target module are tested.}
\label{sample_grids}
\end{figure}

As illustrated in Fig.~\ref{sample_grids}, we define the reference frame $I$ within the horizontal plane, the origin denoted as $O$. The $x-$ and $y-$ axes are aligned with the length and width of the test pool, respectively. The displacement of the target module is denoted as $(p_x,p_y)$, where $p_x$ and $p_y$ represent the displacements along $x-$ and $y-$ axes, respectively. In addition, variable $d$ represents the current direction of the leader vehicle. We assume that the leader vehicle takes one of the three possible orientations, namely, towards the left, moving straight (either forward or backward), and towards the right, denoted as $L$, $S$ and $R$, respectively. Thus we have $d\in\{L,S,R\}$. The states of the leader vehicle, denoted as $\bm{s}$, is represented by the vector $\bm{s}=[p_x,p_y,d]^\text{T}$.

Assume that $P$ acoustic sensors and $Q$ pressure sensors are deployed. Define the measurements of all acoustic sensors at time $k$ as $\bm{m}_k^\text{a}\in\mathbb{R}^P$, and the measurements of all pressure sensors at time $k$ as $\bm{m}_k^\text{p}\in\mathbb{R}^Q$. Additionally, define the 2D camera image captured at time $k$ as $\bm{M}_k^\text{c}\in\mathbb{R}^{U \times V}$, where $U$ and $V$ denote the width and height of the image, respectively. With the multi-modal measurement data, the estimated states of the leader vehicle at time $k$ is calculated through the estimation algorithm function $f$ by
\begin{equation}
\hat{\bm{s}}_{k}=f(\bm{M}_k^\text{c},\bm{m}_k^\text{a},\bm{m}_k^\text{p}),
\end{equation}

At time $k$, define the actual states as $\textbf{\textit{s}}_k$, then the absolute estimation error is calculated by
\begin{equation}
    e_{k} = \left|\hat{\bm{s}}_{k} - \bm{s}_{k}\right|.
\end{equation}

Define the mean Smooth $L_1$ Loss \cite{fastrcnn} as
\begin{equation}
    L_f = \frac{1}{N} \sum_{k=1}^{N} \left[ \frac{1}{2\beta}e_k^2I_{e_k<\beta} + \left(e_k-\frac{\beta}{2}\right)I_{e_k\geq\beta}\right],
\end{equation}
where $N$ is the total quantity of samples, $\beta$ is the threshold hyper-parameter and $I$ is the indicator function.

Finally, the leader state estimation problem is formulated as a task aimed at finding out a function $f^*$ that satisfies
\begin{equation}
     L_{f^*} \leq \epsilon,
\end{equation}
where $\epsilon>0$ is the desired loss limit.
 
\section{Dataset Construction}
This section describes the acquisition process and composition of the multi-modal dataset.

To ensure comprehensive experimentation, we select a set of sample locations. As illustrated in Fig.~\ref{sample_grids}, a mesh grid consisting of 42 locations is chosen, comprising 7 positions along the $x$-axis and 6 positions along the $y$-axis. Along $x$-axis, marginal space is reserved to allow the target module frame to remain stable and flat, while the rear space along the $y$-axis is maintained to distinguish the acoustic measurement threshold.

The multi-modal sensing module is fixed at $(100,50)$ and denoted as \textbf{H}, while the target module, represented by \textbf{T}, is manually placed at the selected locations and remains stationary during data collection.  To simulate the leader vehicle taking different actions, namely turning left, moving straight forward/backward and turning right, the target module is positioned either parallel to the $y$-axis or at angles of $\pm45^\circ$ relative to the $y$-axis. 
Propellers \#1 and  \#2 are defined as illustrated in Fig.~\ref{target}. The rotation speed and rotation direction (RD), either clockwise (CW) or counterclockwise (CCW), of each propeller are controlled by the supply voltage and servo driver's pulse width (PW). During the experiment, the supply voltage is maintained at a constant 15~V with minor variances caused by water flow disturbances. The relationship between target module's actions and the propeller configurations (the servo driver's pulse width PM and the rotation direction RD) is summarized in Table \ref{propeller_action}.
\begin{table}[htpb]
\vspace{-10pt}
\caption{Propeller Configurations for Target Module's Actions}
\vspace{-5pt}
\begin{center}
\renewcommand\arraystretch{1.4}
\begin{tabular}{ccccc}
\toprule
\multirow{2}*{Action} & \multicolumn{2}{c}{Propeller \#1} & \multicolumn{2}{c}{Propeller \#2}\\
\cmidrule(lr){2-3}\cmidrule(lr){4-5}&
PW (\textmu s) & RD & 
PW (\textmu s) & RD \\
\midrule
Move Straight& 2K & CCW & 1K & CW\\
Turn Left & 1K & CW & 1K & CW\\
Turn Right & 2K & CCW & 2K & CCW\\
\bottomrule
\end{tabular}
\end{center}
\label{propeller_action}
\end{table}

At each of the 42 predefined locations,  multi-modal sensory data are collected with the target module oriented in three distinct directions. This results in a total of 126 sampled cases. To ensure clarity in the experiment description, the term \textit{location} refers to one of the 42 spatial points, while the term \textit{case} denotes one of the 126 combinations of target module's locations and orientations.

In the following sections, a case is represented by the target module's state $\bm{s}$ in the triplet form $\left(p_x, p_y, d\right)$. For example, the case (40,70,$L$) refers to the scenario where the propeller executes a left-turn action at the location (40,70).
\begin{figure}[thbp]
\centering
\subfigure[Case (60,90,$M$)]{
\label{img_090060}
\includegraphics[width=0.15\textwidth]{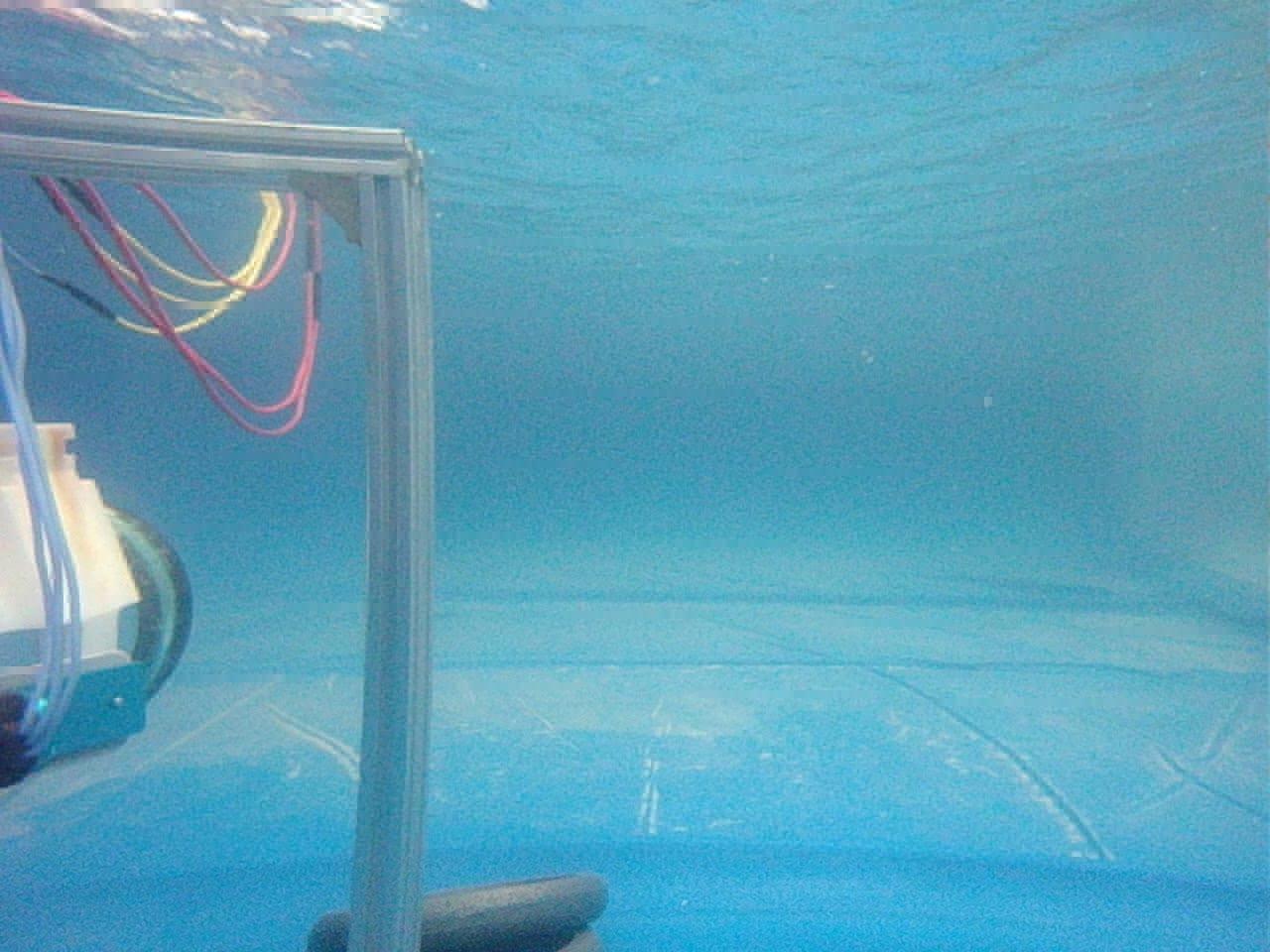}}
\subfigure[Case (60,190,$M$)]{
\label{img_190060}
\includegraphics[width=0.15\textwidth]{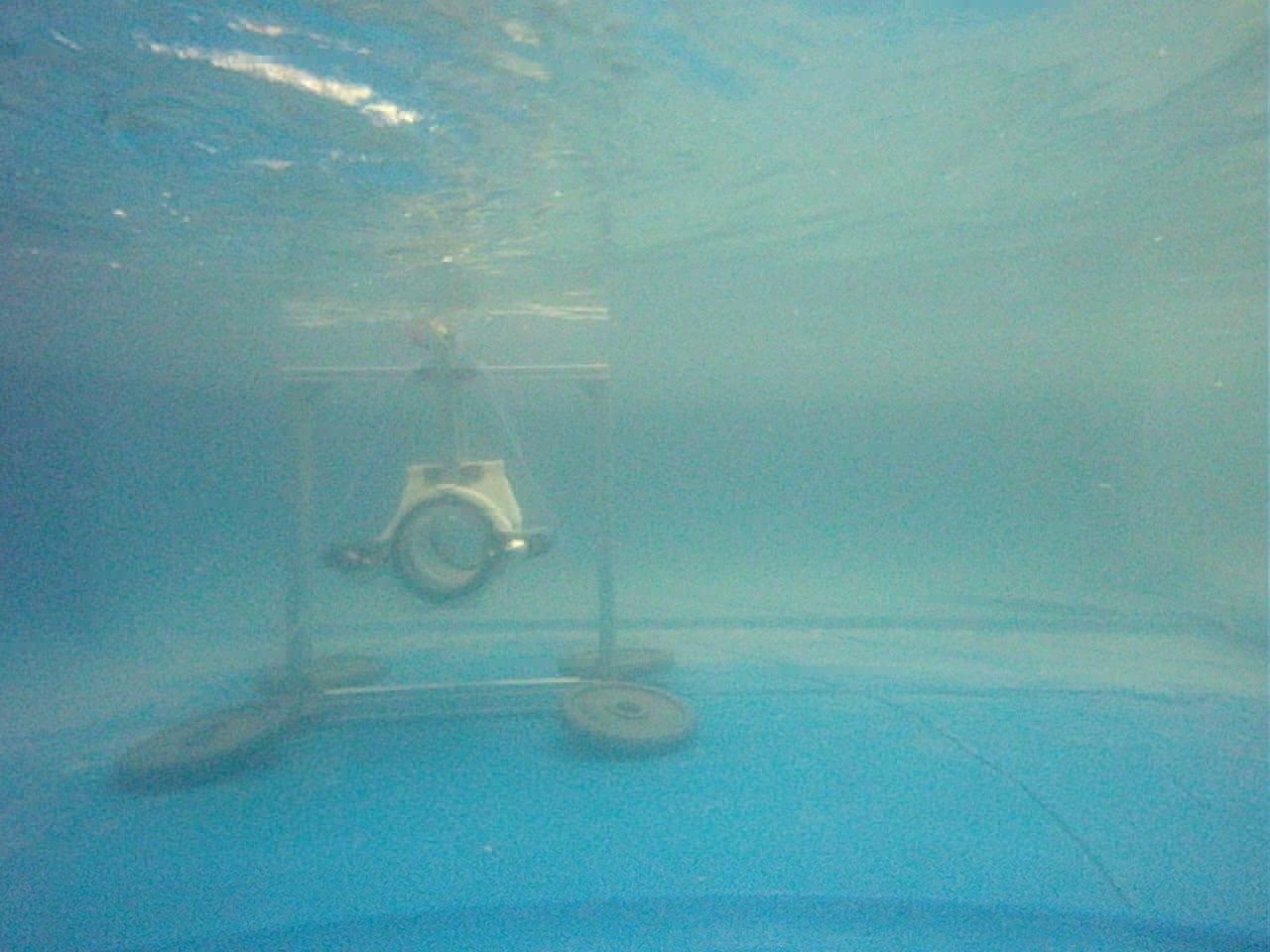}}
\subfigure[Case (60,290,$M$)]{
\label{img_290060}
\includegraphics[width=0.15\textwidth]{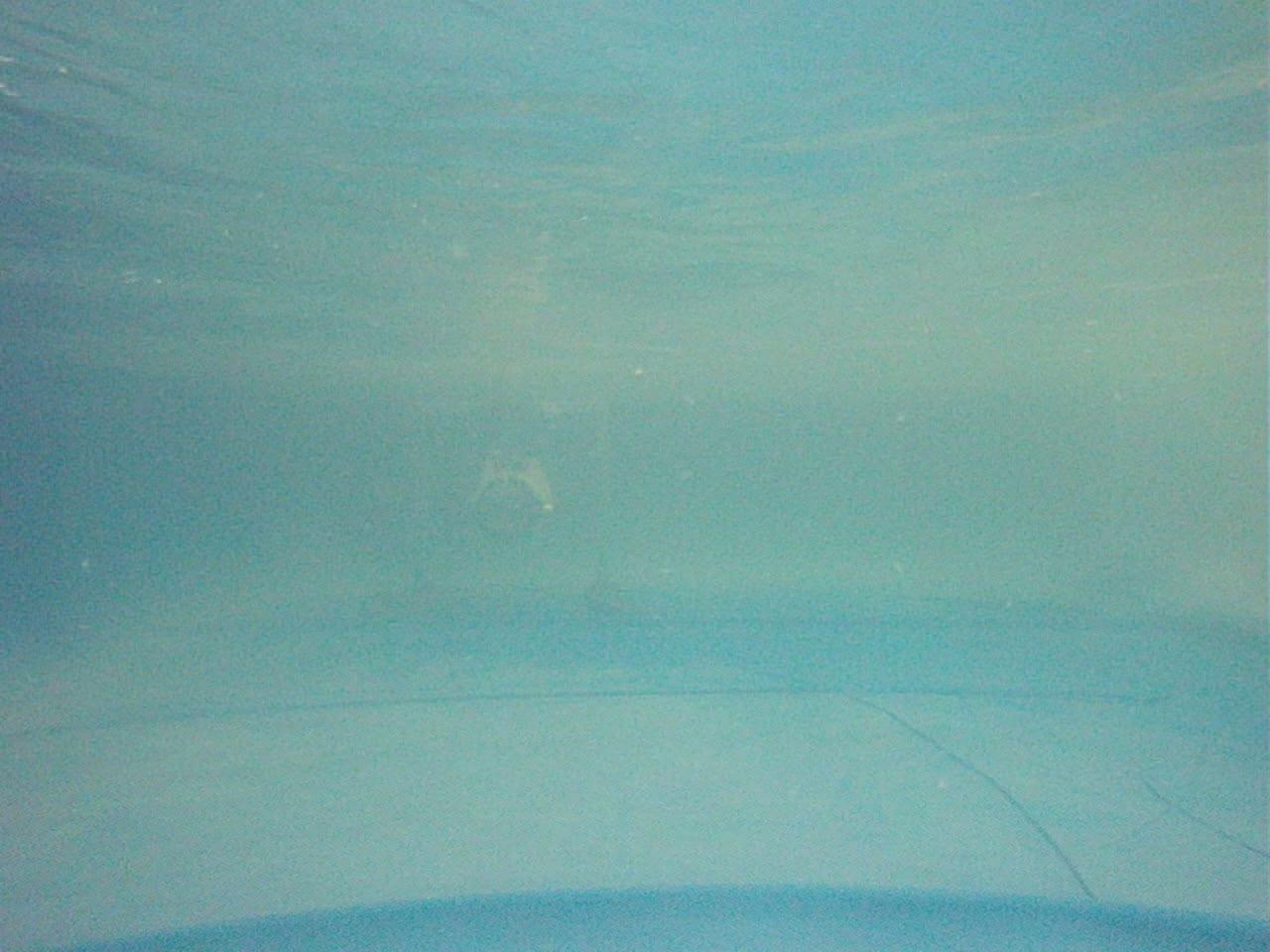}}
\subfigure[Case (100,90,$M$)]{
\label{img_090100}
\includegraphics[width=0.15\textwidth]{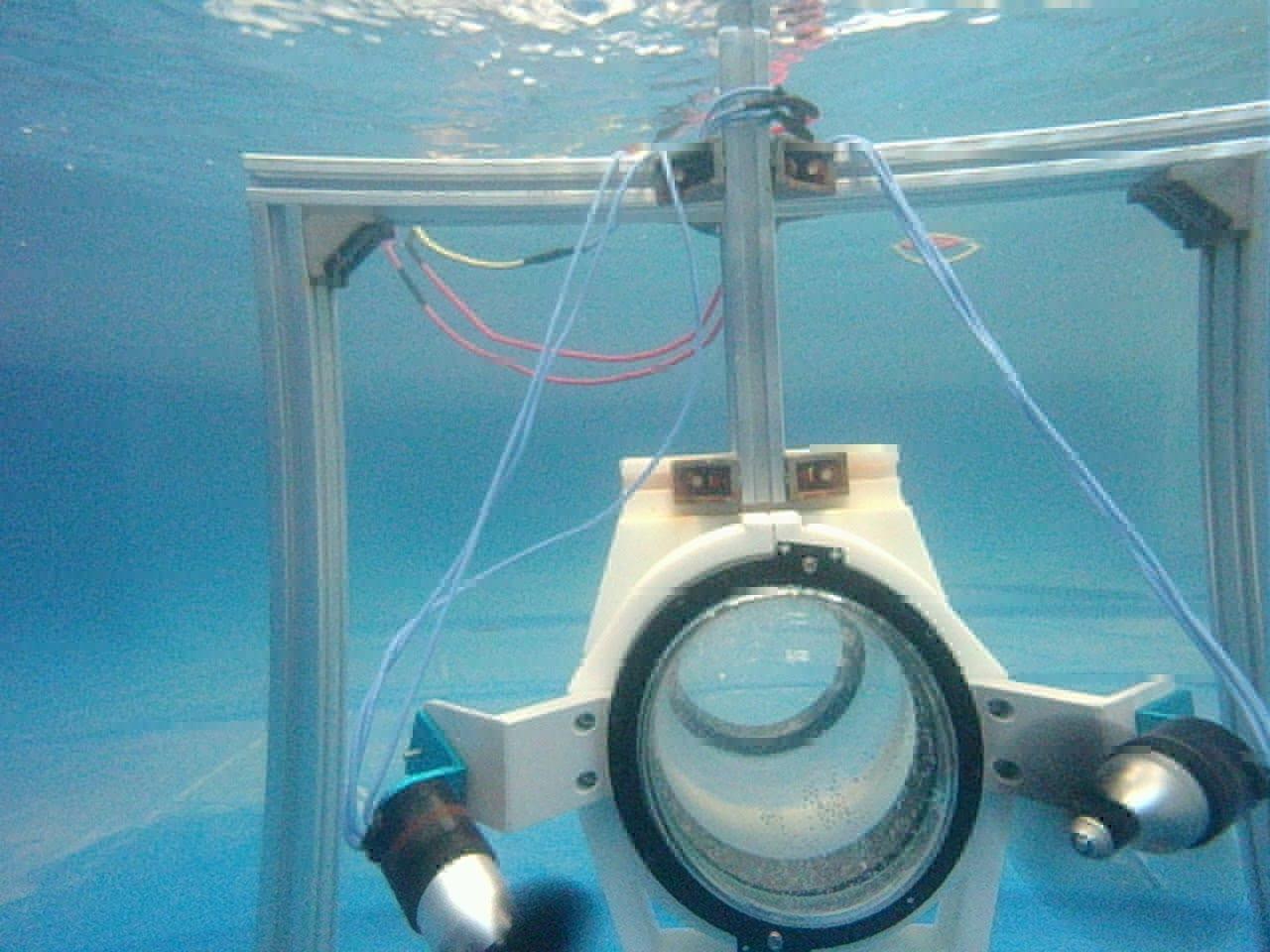}}
\subfigure[Case (100,190,$M$)]{
\label{img_190100}
\includegraphics[width=0.15\textwidth]{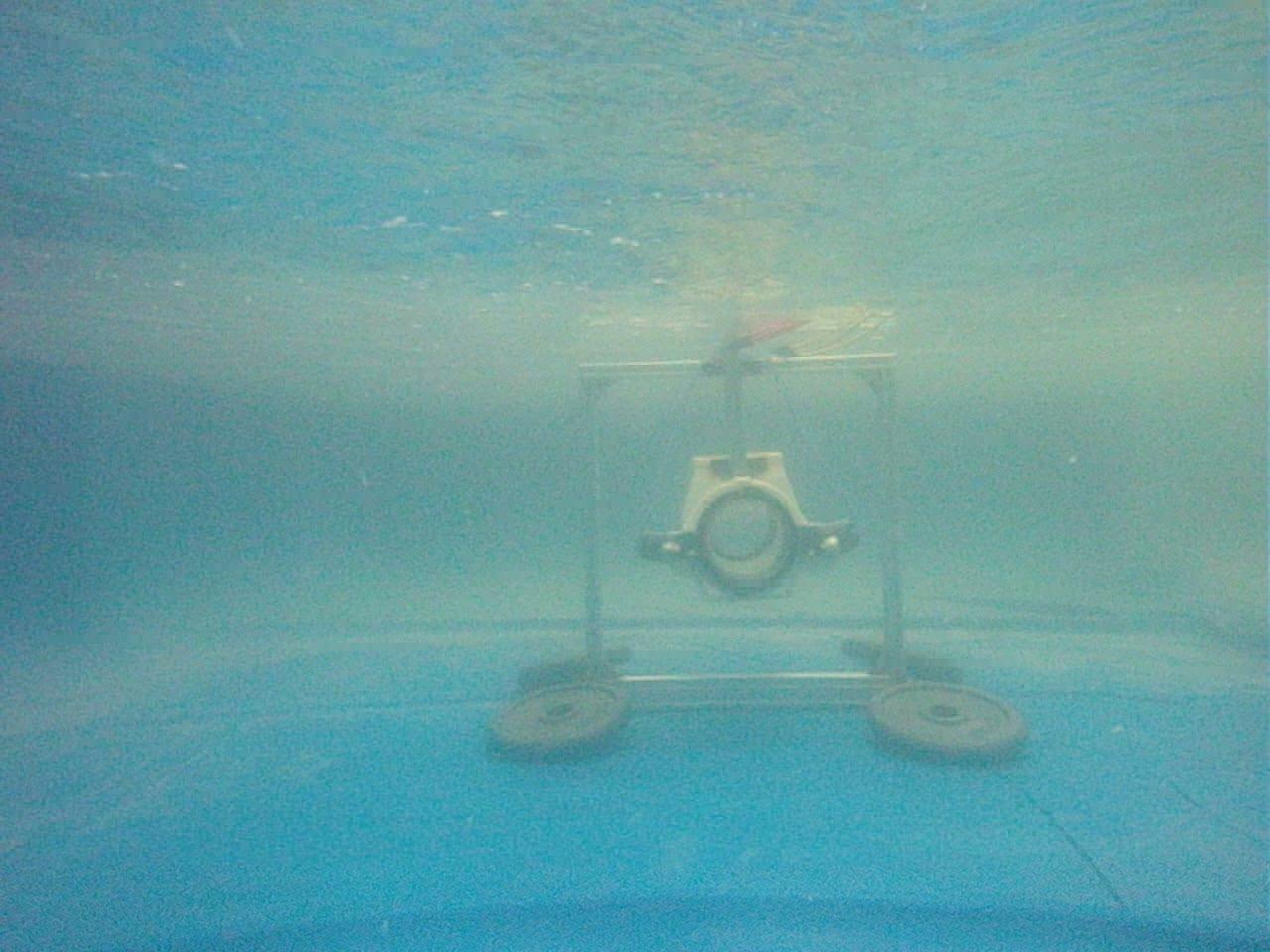}}
\subfigure[Case (100,290,$M$)]{
\label{img_290100}
\includegraphics[width=0.15\textwidth]{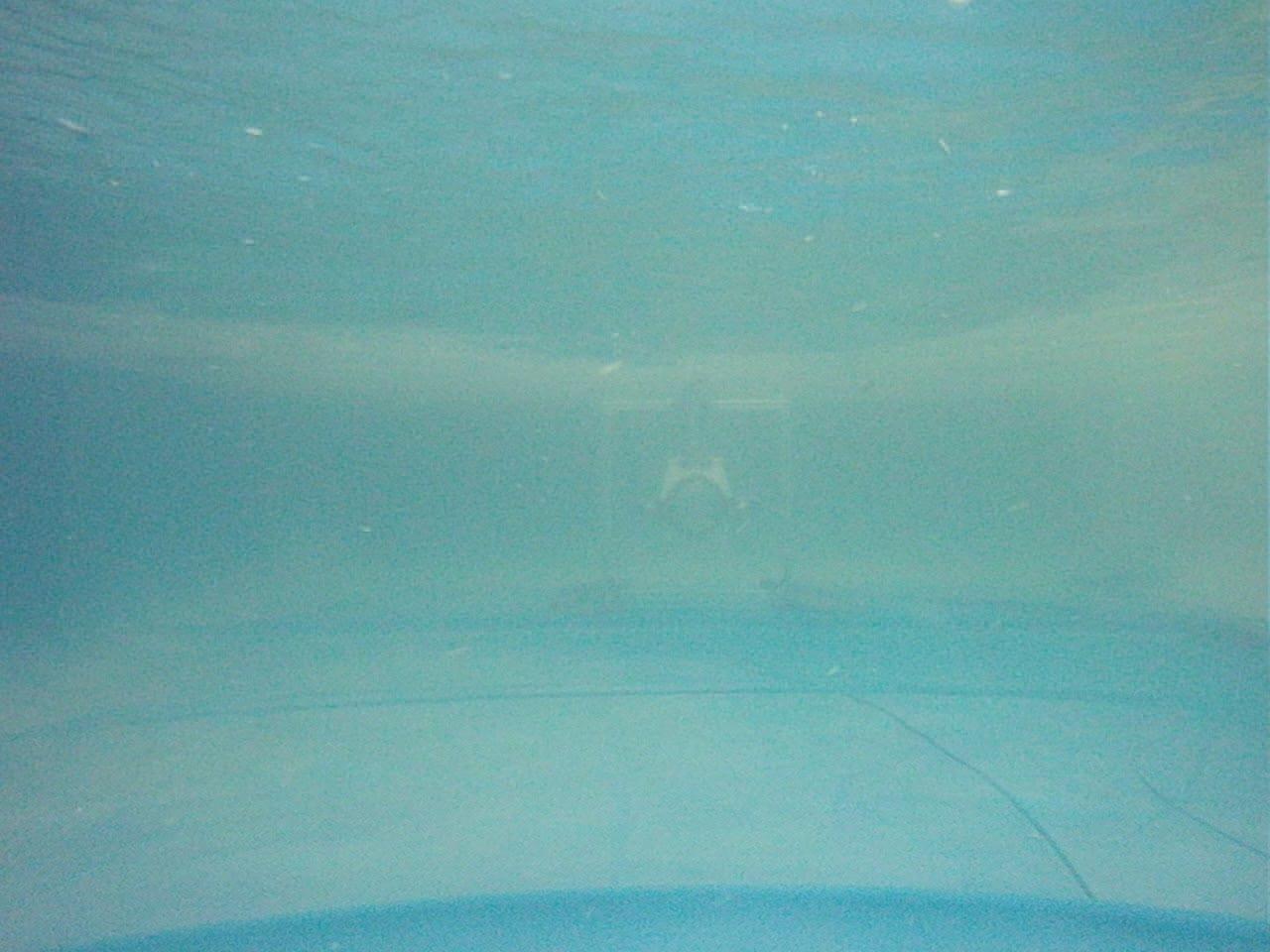}}
\caption{Visualization of the collected raw image data for the provided case triplets. At $p_y=90$~cm, the target appears crystal clear but only partially within the camera field of view (FOV). At $p_y=190$~cm, the target is mostly within the camera FOV but less visible. At $p_y=290$~cm, the target is fully within in the camera FOV but barely visible due to increased distance and water turbidity.}
\label{img_data_vis}
\end{figure}

\textit{Visual Data Collection:} The camera operates at a sample frequency of 4~fps. For each case, a 12-minute video is recorded. Fig.~\ref{img_data_vis} visualizes several representative raw images from  the collected dataset, annotated with their corresponding case triplets.

\begin{figure}[thpb]
 \centering
 \includegraphics[width=0.45\textwidth]{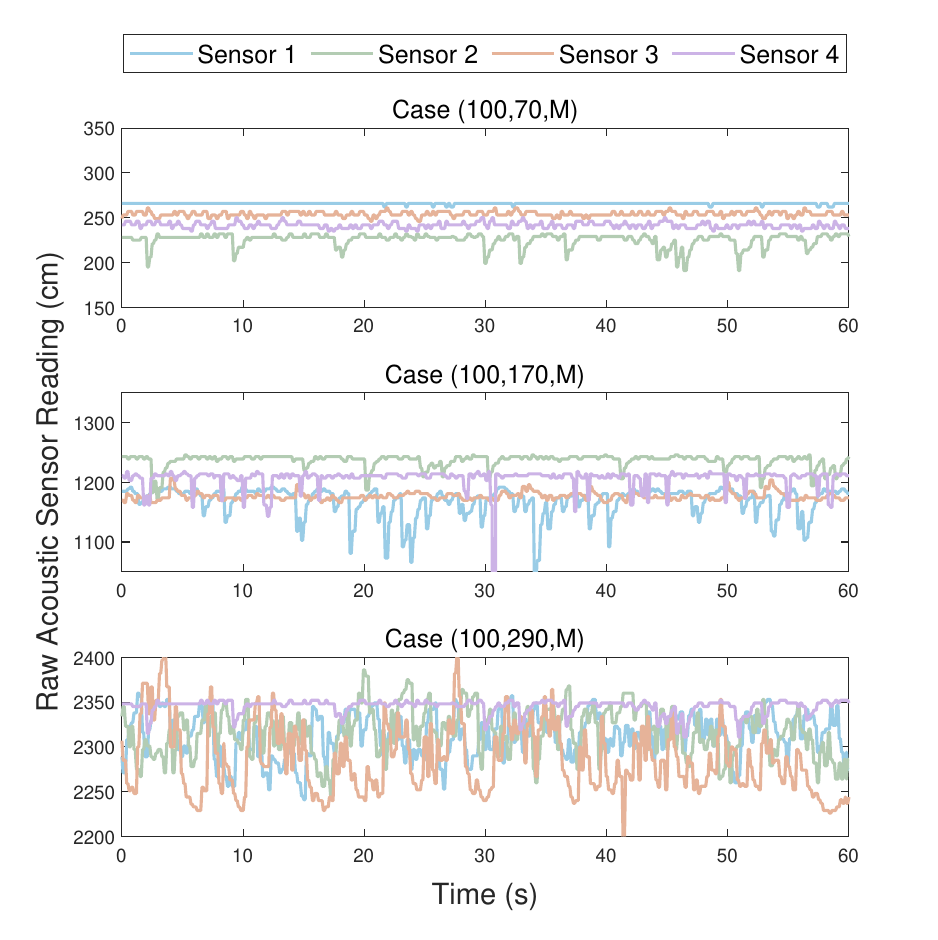}
 \vspace{-5pt}
 \caption{Raw acoustic measurements from four ultrasonic ranging sensors over a 60-second duration for six cases. The case triplets are indicated above each curve plot. When the target is closer to the sensing module, the variance of the readings from a specific acoustic sensor is significantly smaller.}
\label{acoustic_raw_vis}
\end{figure}

\textit{Acoustic Data Collection:} Four ultrasonic ranging sensors are employed. To mitigate cross-talk issues, the sensors operate  sequentially at a sample frequency of 10~Hz each. Fig.~\ref{acoustic_raw_vis} illustrates raw acoustic measurements for 3~cases.

\begin{figure}[thpb]
 \centering
 \includegraphics[width=0.45\textwidth]{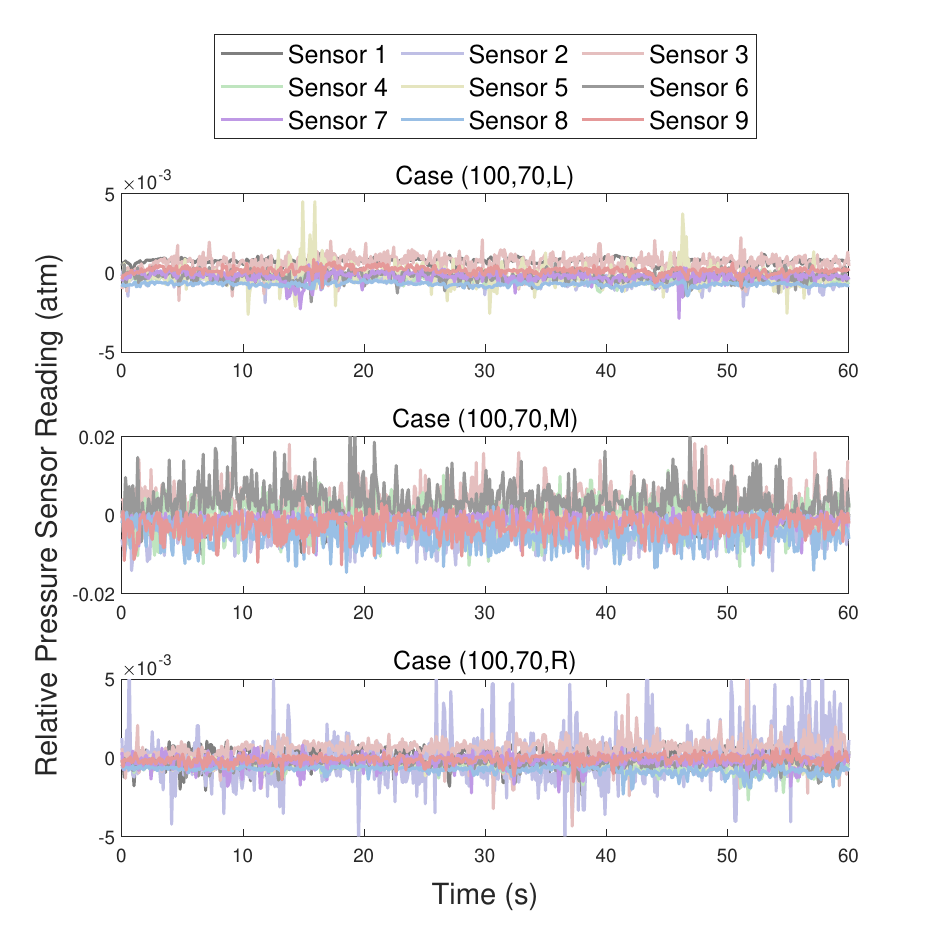}
 \vspace{-5pt}
 \caption{Relative pressure measurements from nine pressure sensors over a 60-second duration. The case triplets are indicated above each curve plot.}
\label{pressure_relative_vis}
\end{figure}

\textit{Pressure Data Collection:} Nine distributed pressure sensors are used to capture the propellers' wake flow. For each case, the nine pressure sensors collaboratively sample the wake generated by the propellers at the frequency of 10~Hz. Prior to propeller operation, still water pressures are recorded for 30~seconds for each case. For each of the 126 cases, denoted as $c$, 
and sensor $i\in\left[1, 2, 3, \ldots, 9 \right]$, the recorded still water measurements are represented by $\left[r_{ci}^1, r_{ci}^2, ..., r_{ci}^{sl}\right]$. The still water pressure for each case and sensor is calculated by $r^\text{s}_{ci}=\frac{1}{L}\sum_{k=1}^{sl}r^k_{ci}$,
where $sl$ is the sequence length of the still water measurements. At time $k$, the relative pressure sensor reading is calculated by $r^{k'}_{ci}=r^k_{ci}-r^\text{s}_{ci}$.  Fig.~\ref{pressure_relative_vis} illustrates the relative pressure sensor readings for six 6~cases, measured in atmospheres (atm). Each illustrated case spans 60~seconds. 
During preliminary experiments, we observe that the pressure sensors are highly effective in detecting the propeller wake when the target module is closer to the sensing module (e.g., $p_y \leq 100$). However, as the distance increase (e.g., $p_y \geq 100$), the fluctuations in pressure readings became indistinguishable from background noise. This limitation is inherent to the physics of pressure wave propagation in water, where signal strength diminishes with distance. Consequently,  pressure sensor data are utilized for target module localization only when $p_y\in\{70,90\}$.

\section{End-to-end State Estimation Algorithm}
This section presents a novel end-to-end neural network architecture that fuses three sensing modalities, i.e., visual, acoustic and pressure, to estimate the states of a leader vehicle $\hat{\bm{s}}_k$.

\begin{figure*}[thpb]
 \centering
 \includegraphics[width=1\textwidth]{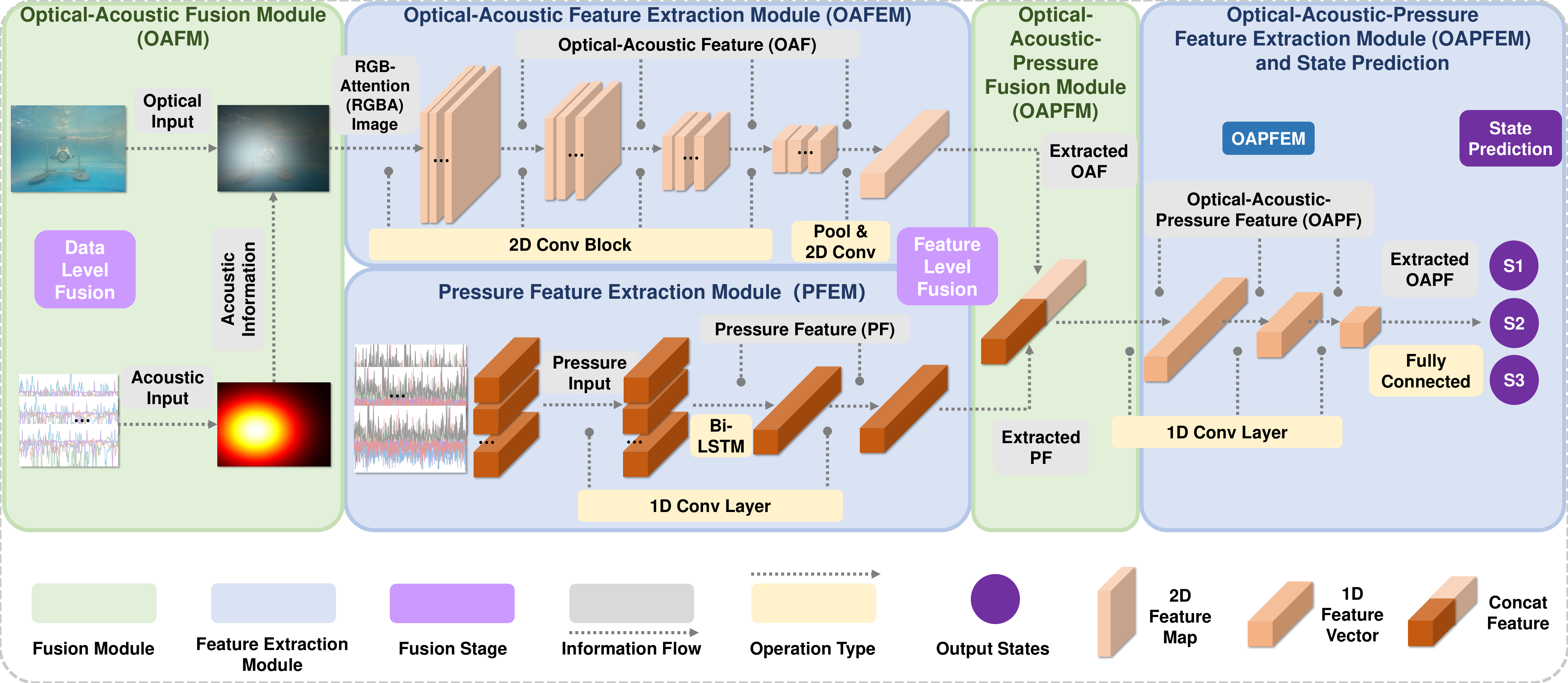}
 \caption{Overview of the end-to-end algorithm architecture for leader vehicle localization. The optical images and the acoustic ranging measurements are fused at  the data level (RGBA Image) within the Optical-Acoustic Fusion Module (OAFM). Subsequently, the Optical-Acoustic Feature Extraction Module (OAFEM) applies a sequence of convolutional networks to extract optical-acoustic features (OAF).  Pressure sensor measurements are processed through convolutional and recurrent networks in the Pressure Feature Extraction Module (PFEM) to extract pressure features (PF). The extracted OAF and PF are fused at the feature level in the Optical-Acoustic-Pressure Fusion Module (OAPFM). The combined information from all three modalities is processed fully-connected networks in the Optical-Acoustic-Pressure Feature Extraction Module (OAPFEM) to extract the fused optical-acoustic-pressure feature (OAPF), which is used in the final state estimation of the target module.}
\label{Algflow}
\end{figure*}

The overall design is illustrated in Fig.~\ref{Algflow}. It comprises five modules including two information fusion modules (highlighted in green) and three feature extraction modules (highlighted in blue). Optical-Acoustic Fusion Module (OAFM) performs data-level fusion of acoustic and optical information. The resulting fused optical-acoustic data is processed in the Optical-Acoustic Feature Extraction Module (OAFEM) to extract the optical-acoustic feature (OAF). This feature is subsequently fused with the pressure feature (PF) at the feature level in Optical-Acoustic-Pressure Fusion Module (OAPFM), following processing by the Pressure Feature Extraction Module (PFEM), which outputs the PF. The fused optical-acoustic-pressure feature (OAPF) is then processed by the Optical-Acoustic-Pressure Feature Extraction Module (OAPFEM) to extract the integrated tri-modal feature. The leader state estimation is finally performed based on this tri-modal feature. 

\subsection{Optical-Acoustic Information Fusion (OAFM)}
\label{OAFM}
In OAFM, optical information and acoustic information are fused at data level through heatmap attention mechanism.

For most natural images, visual information is stored across the Red, Green and Blue (RGB) channels. However, in certain cases, a fourth channel, the Alpha channel, is supported, resulting in an RGBA image. The transformation from an RGBA image to an RGB image is governed by
\begin{equation}
    v_j^{\text{rgb}} = v_j^{\text{rgba}}\times v_\alpha + v_j^\text{b}\times(1-v_\alpha),
\end{equation}
where $j\in$\{R,G,B\}, $v_j^\text{rgb}$ represents the channel value in the RGB image, $v_j^\text{rgba}$ represents the corresponding channel value in the RGBA image, $v_\alpha$ denotes Alpha channel value, and $v_j^\text{b}$ represents background RGB value.

Inspired by the RGBA color model, the OAFM incorporates acoustic information by overlaying it as a heatmap attention channel on the camera's RGB image, thereby forming an RGB-Attention image, abbreviated as an RGBA image. In the following sections, RGBA image refers to a fused image combining visual (RGB) data with an ``attention channel" derived from acoustic sensor data. The attention channel highlights regions of interest to guide the neural network.

The fusion process proceeds as follows. 
First, \textit{Data Preparation.} Acoustic ranging sensor data is filtered using upper and lower thresholds.  
   The mean ($\mu_\text{a}$) and standard deviation ($\sigma_\text{a}$) of the filtered data are calculated,  
   and training/testing data, \(r_\text{a}\), is generated via a Gaussian distribution, i.e., $r_\text{a}\sim N\left(\mu_\text{a},\sigma_\text{a}^2\right)$.
Second, \textit{Heatmap Generation.} Each acoustic sensor's  transmission and reception fields are modeled as a cone beam \cite{qiu2022review}, with uncertainty represented by Gaussian heatmap. Sensor positions relative to the camera are defined by translation vectors $t_u^j$ and $t_v^j$. Using  $r_\text{a}$, boundaries in the image frame ($u_\text{u}, u_\text{l}, v_\text{u}, v_\text{l}$) are determined via pinhole camera model transformation $g$. 
Third, \textit{Heatmap Attention.} For each acoustic sensor, the mean vector $\bm{\mu}^j=\left(\mu_u^j,\mu_v^j\right)^\text{T}$ and covariance matrix $\bm{C}^j=$ diag$({\sigma_u^j}^2,{\sigma_v^j}^2)$ 
are calculated by
$\mu_u^j = (u_\text{u}+u_\text{l})/{2}$, 
$\mu_v^j = (v_\text{u}+v_\text{l})/{2}$,
$\sigma_u^j = (u_\text{u} - u_\text{l}) /{\gamma_u}$, 
$\sigma_v^j = (v_\text{u} - v_\text{l}) / {\gamma_v}$,
where $\left(\gamma_u,\gamma_v\right)$ are the expansion factors used to control the attention significance.
Fourth, \textit{Joint Heatmap Fusion.} If the target module is within the reception fields of $J$ acoustic sensors, where $2\leq J\leq4$, individual heatmaps are fused to create a joint heatmap. The joint mean $\bm{\mu}^\text{J}=\left(\mu_u^\text{J},\mu_v^\text{J}\right)^\text{T}$ and joint covariance $\bm{C}^\text{J}=$ diag$({\sigma_u^\text{J}}^2,{\sigma_v^\text{J}}^2)$ are calculated by
\begin{equation}
    \mu_u^\text{J} = \frac{1}{J}\sum_{j=1}^J\mu_u^j, \quad
    \sigma_u^J=\sqrt{\sum_{j=1}^J(\sigma_u^\text{j})^2}
\end{equation}
with similar equations for $\mu_v^\text{J}$ and $\sigma_v^J$. The fused heatmap is  normalized to $\left[0,1\right]$.
Last, \textit{RGBA Image Formation.}
The joint acoustic heatmap is concatenated with the corresponding RGB image as an attention channel, forming the RGBA optical-acoustic image.

\begin{figure}[thpb]
\centering
\setlength{\belowcaptionskip}{-15pt}
\subfigure{
\hspace{-3mm}
\rotatebox{90}{\scriptsize{~~~Case (80,90,R)}}
\begin{minipage}[t]{0.15\textwidth}
\centering
\includegraphics[width=1\textwidth]{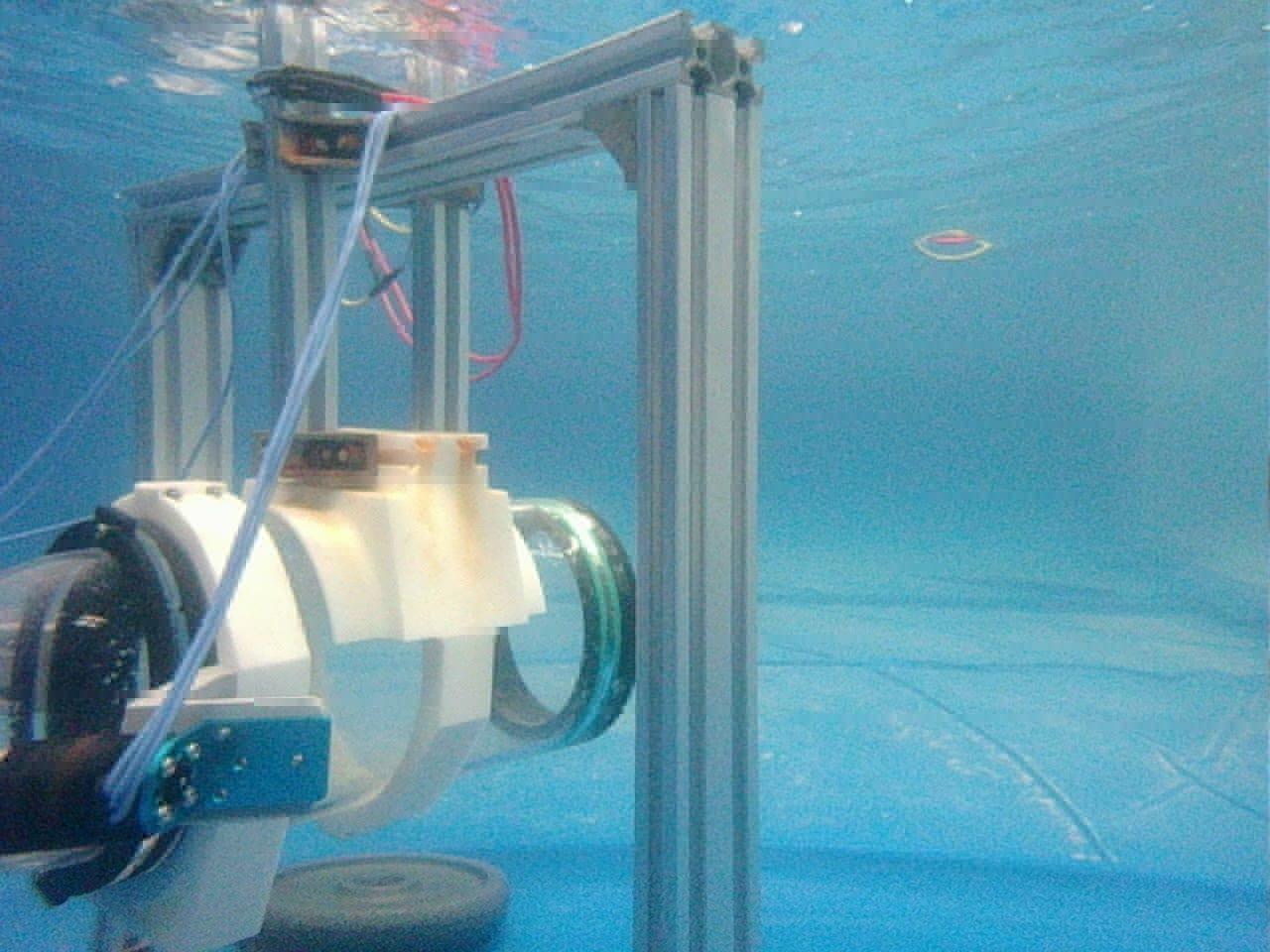}
\end{minipage}}
\hspace{-3mm}
\subfigure{
\begin{minipage}[t]{0.15\textwidth}
\centering
\includegraphics[width=1\textwidth]{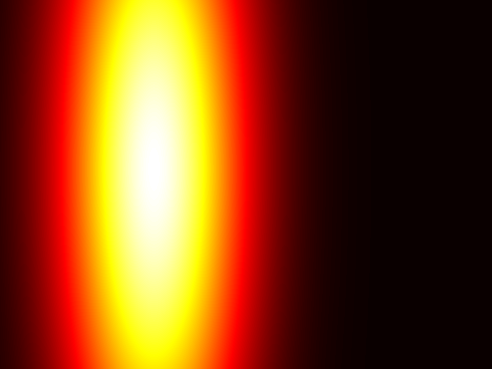}
\end{minipage}}
\hspace{-3mm}
\vspace{-2.6mm}
\subfigure{
\begin{minipage}[t]{0.15\textwidth}
\centering
\includegraphics[width=1\textwidth]{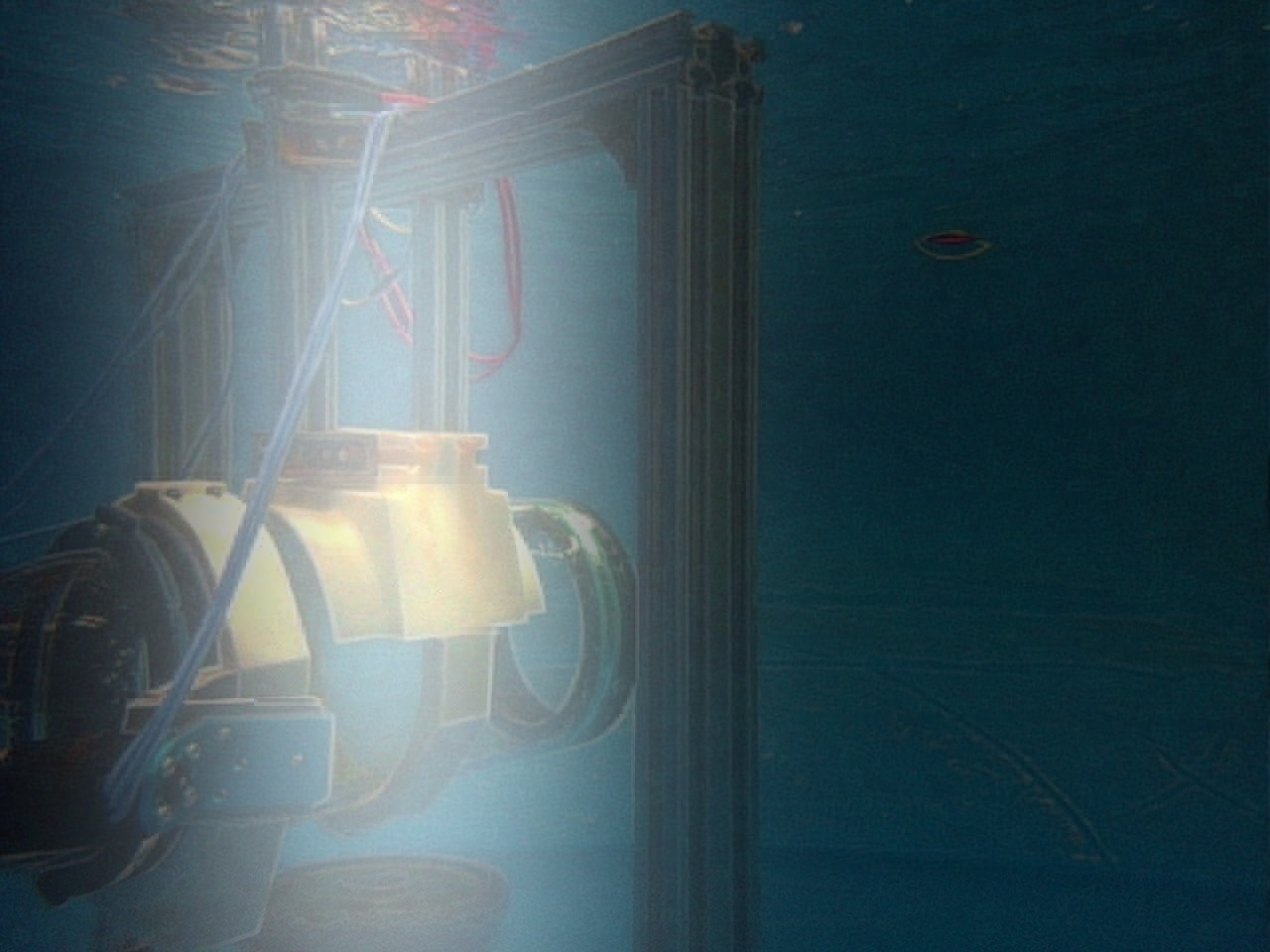}
\end{minipage}}
\subfigure{
\hspace{-3mm}
\rotatebox{90}{\scriptsize{~Case (140,190,M)}}
\begin{minipage}[t]{0.15\textwidth}
\centering
\includegraphics[width=1\textwidth]{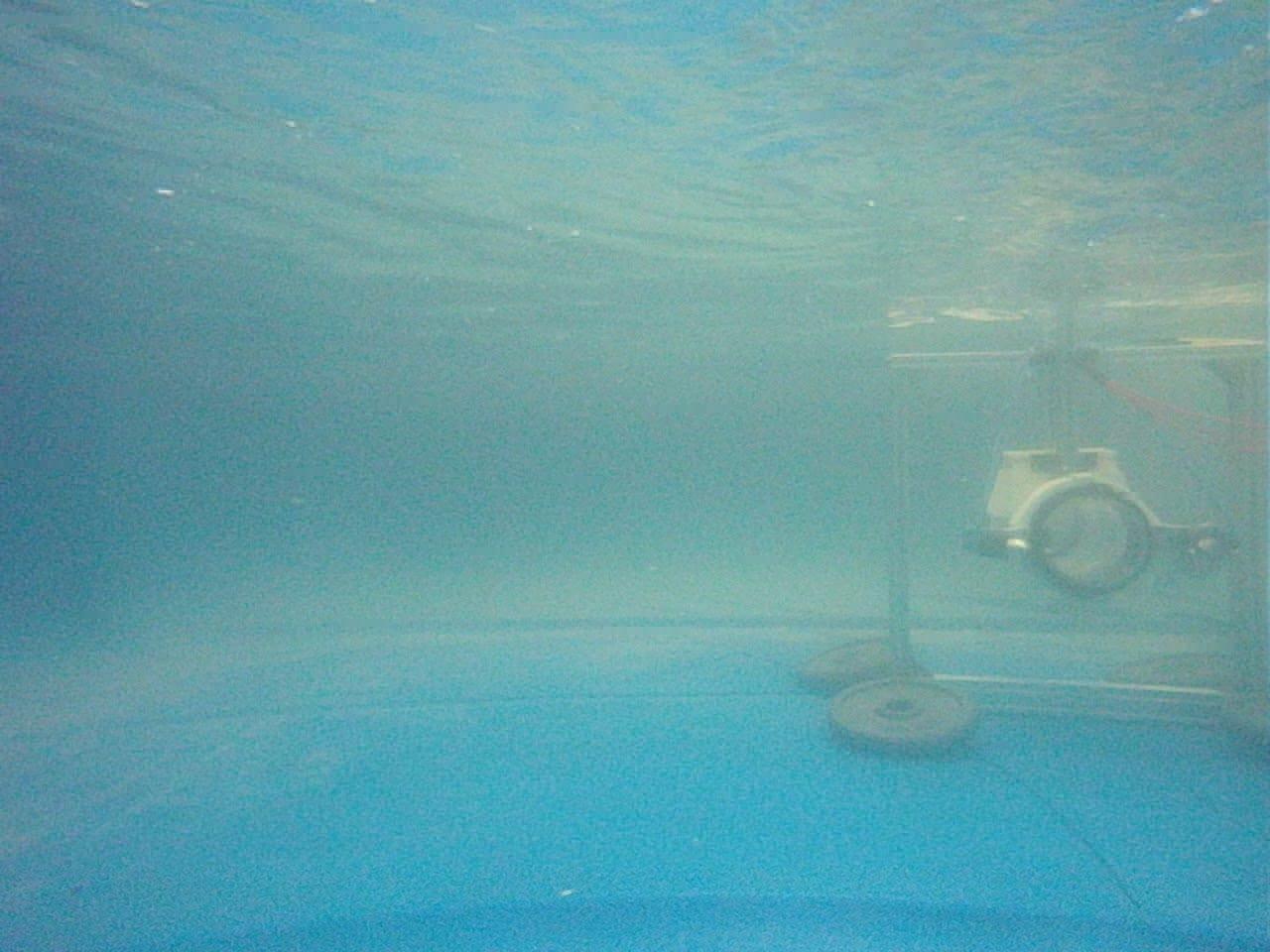}
\end{minipage}}
\hspace{-3mm}
\subfigure{
\begin{minipage}[t]{0.15\textwidth}
\centering
\includegraphics[width=1\textwidth]{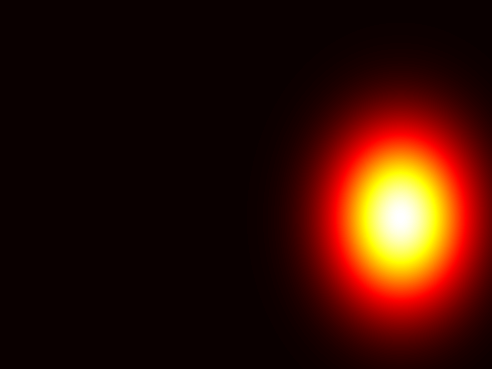}
\end{minipage}}
\hspace{-3mm}
\vspace{-2.7mm}
\subfigure{
\begin{minipage}[t]{0.15\textwidth}
\centering
\includegraphics[width=1\textwidth]{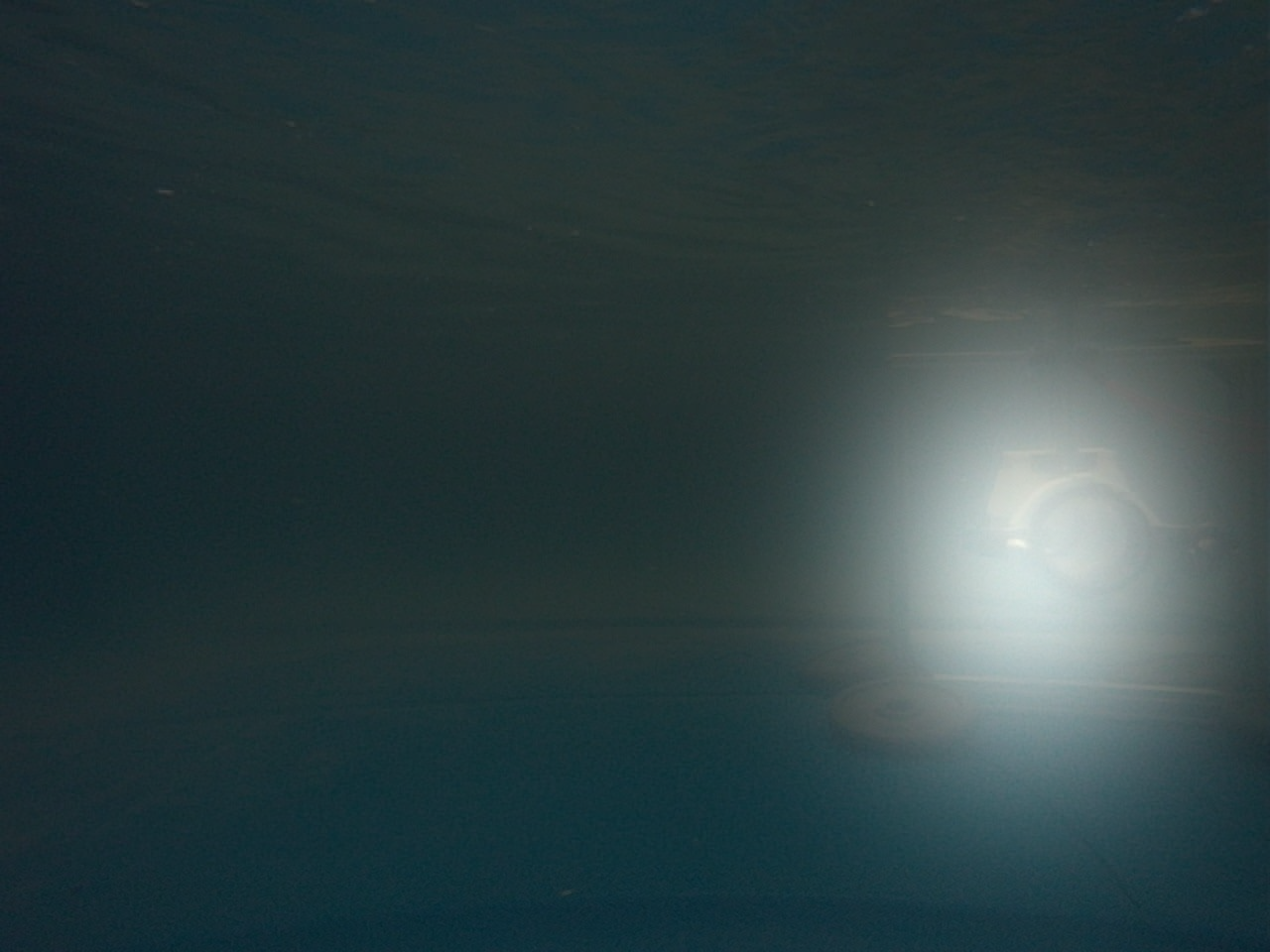}
\end{minipage}}
\setcounter{subfigure}{0}
\subfigure[Camera RGB image.]{
\hspace{-3mm}
\rotatebox{90}{\scriptsize{~~Case (60,290,M)}}
\begin{minipage}[t]{0.15\textwidth}
\centering
\includegraphics[width=1\textwidth]{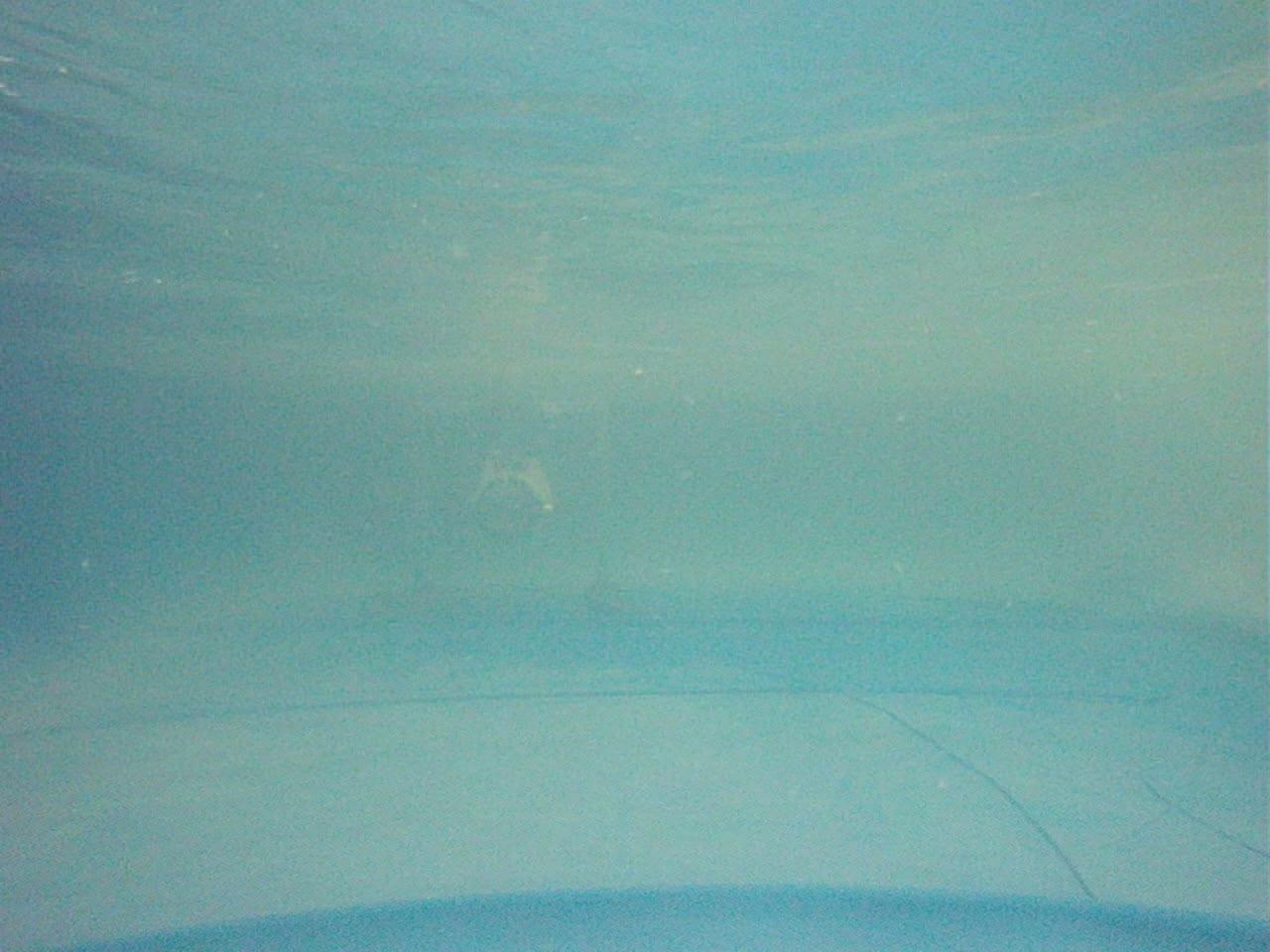}
\label{RGB}
\end{minipage}}
\hspace{-3mm}
\subfigure[Acoustic heatmap.]{
\begin{minipage}[t]{0.15\textwidth}
\centering
\includegraphics[width=1\textwidth]{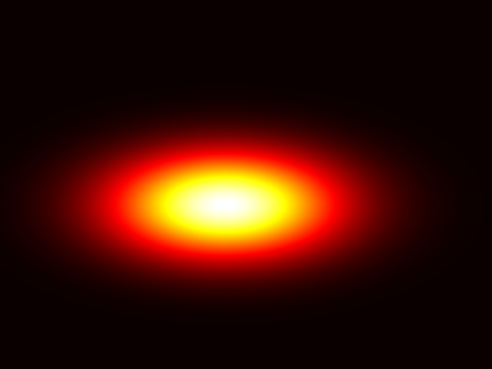}
\label{acoustic_hm}
\end{minipage}}
\hspace{-3mm}
\subfigure[Fused RGBA image.]{
\begin{minipage}[t]{0.15\textwidth}
\centering
\includegraphics[width=1\textwidth]{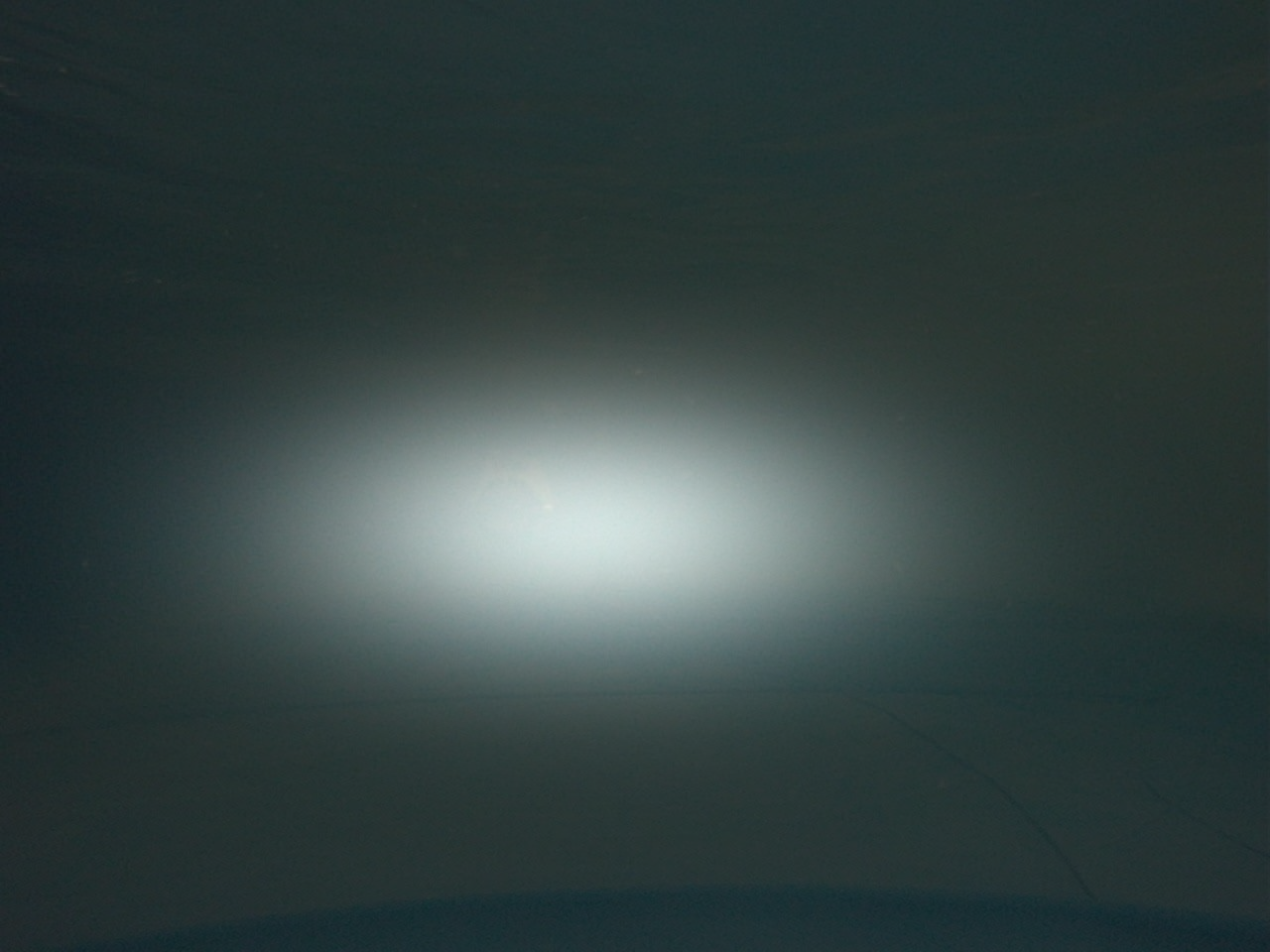}
\label{RGBA}
\end{minipage}}
\caption{Visualization of three optical-acoustic fusion examples. Each row represents one of three cases. The leftmost column displays the original camera RGB image, the middle column displays the generated joint acoustic heatmap and the rightmost column displays the fused RGBA image, respectively.}
\label{optical-acoustic-fusion}
\end{figure}

Three examples with RGB image, generated joint acoustic heatmap, and fused RGBA image for each example are illustrated in Fig.~\ref{RGB}, Fig.~\ref{acoustic_hm}, and Fig.~\ref{RGBA}, respectively.

\subsection{Optical-Acoustic Feature Extraction (OAFEM)}
In OAFEM, G-GhostNet \cite{ghostnetv1,gghostnet}, a lightweight convolutional neural network architecture designed to reduce computational costs while maintaining accuracy by generating ``ghost features" from inexpensive operations, is adopted as the backbone structure. 
We modified the backbone to suit the OAFEM module's requirements, as detailed in Table~\ref{g-ghost}, where Block denotes the residual bottleneck specified in \cite{gghostnet} and \#Out represents the number of output channels.
\begin{table}[thpb]
\caption{Architecture of the OAFEM}
\begin{center}
\renewcommand\arraystretch{1.4}
\begin{tabular}{cccc}
\toprule
Stage & Output Size & Operator & \#Out\\
\midrule
stem & 112$\times$112 & Conv$\left(3,3\right)$ & 16\\
\midrule
1 & 56$\times$56 & Block, Block$\times$1 Cheap, Concat & 24\\
\midrule
2 & 28$\times$28 & Block, Block$\times$3 Cheap, Concat & 48\\
\midrule
3 & 14$\times$14 & Block, Block$\times$3 Cheap, Concat & 96\\
\midrule
4 & 7$\times$7 & Block, Block$\times$5 Cheap, Concat & 192\\
\midrule
\multirow{2}*{5} & 7$\times$7 & Conv$\left(1,1\right)$ & 512\\
 & 1$\times$1 & Pool \& Conv$\left(1,1\right)$, Conv$\left(1,1\right)$ & 256\\
\bottomrule
\end{tabular}
\end{center}
\label{g-ghost}
\end{table}

\subsection{Pressure Feature Extraction (PFEM)}
PFEM is designed to capture the spatio-temporal dependencies of the nine pressure sensor measurements. A hybrid architecture combining 1D Convolutional Neural Network (CNN) and Bidirectional Long Short-Term Memory (BiLSTM), a type of recurrent neural network that processes sequential data in both forward and backward directions, capturing long-range temporal dependencies, is employed, enabling the module to effectively extract complex spatiotemporal features of the propeller wake. The PFEM comprises four sequential 1D CNN layers followed a BiLSTM layer.

\subsection{Optical-Acoustic-Pressure Information Fusion (OAPFM) and Optical-Acoustic-Pressure Feature Extraction (OAPFEM)}
In OAPFM, features are concatenated along the channel dimension. Subsequently, OAPFEM employs a series of convolutional layers followed by a fully connected (FC) layer to process the fused features for the final estimation.

\subsection{State Estimator}
\label{SE}
The network outputs three regression values within the range $\left[0,1\right]$, representing the $x$-axis position $p_x$, the $y$-axis position $p_y$, and vehicle direction $d$.
\begin{figure*}[t]
 \centering
 \includegraphics[width=0.9\textwidth]{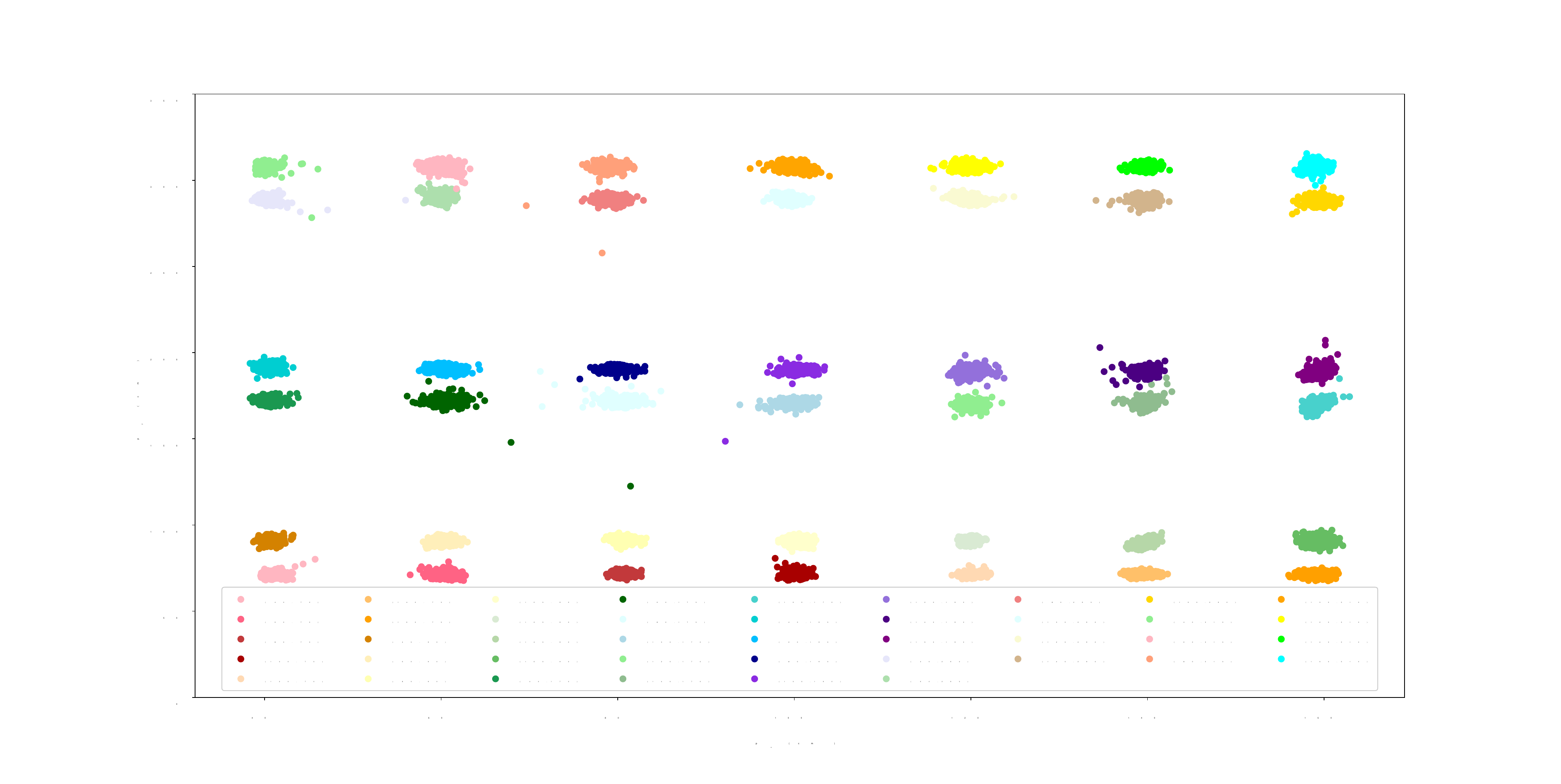}
 \caption{Visualization of position states $(p_x, p_y)$ in a scatter plot, excluding the direction state. The horizontal and vertical axes represent the $x$-axis and $y$-axis, respectively, as shown in Fig.~\ref{sample_grids}. Each color corresponds to a location, with each circle representing the estimation result for a test instance. For locations where pressure data is available as an additional information source (i.e., $p_y \in \{70, 90\}$), the variance in the estimates with the same ground truth is smaller compared to locations with $p_y \in \{170, 190, 290, 310\}$, indicating a more accurate and robust state estimation process.}
\label{avp_pos_vis}
\end{figure*}

As illustrated in Fig.~\ref{sample_grids}, the absolute $x$ and $y$ coordinates of the sampled locations are $[40, 60, 80, 100, 120, 140, 160]$~cm and $[70, 90, 170, 190, 290, 310]$~cm, respectively. The position regression values are normalized to [0,1] using $p_\text{n} = \left(p-p_\text{min}\right)/\left(p_\text{max}-p_\text{min}\right)$ with $p_x\in\left[0,200\right]$~cm and  $p_y\in\left[0,400\right]$~cm. The normalized $x-$ and $y-$ coordinates, represented by $p_{x\text{n}}$ and $p_{y\text{n}}$, are $[0.2, 0.3, 0.4, 0.5, 0.6, 0.7 ,0.8]$ and $[0.175, 0.225, 0.425, 0.475, 0.725, 0.775]$, respectively. Vehicle direction values, turning left, moving straight, and turning right are represented by 0, 0.5 and 1, respectively.

A smooth $L_1$ loss function is adopted for these three states, represented by $l_\text{x}, l_\text{y}$ and $l_\text{d}$. The total estimation loss is calculated by $l_\text{t}=\lambda_\text{x}l_\text{x}+\lambda_\text{y}l_\text{y}+\lambda_\text{d}l_\text{d}$, where $\lambda_\text{x}, \lambda_\text{y}$ and $\lambda_\text{d}$ are weight factors.

\section{Experiments}

\subsection{Training Environment and Hyper-parameters}
In the training process, the batch size is set to 128, with a maximum of 200 epochs. The learning rate is set to 0.1 initially and decays by a fctor of 10 every 30 epochs. Stochastic Gradient Descent (SGD) is adopted for optimization. To mitigate overfitting, weight decay and dropout techniques are applied. The training and testing processes are implemented on a desktop computer equipped with NVIDIA GeForce RTX 3070 GPU and an Intel Core i7-12700F CPU.

\subsection{Results and Discussion}
To evaluate the effectiveness of the proposed leader state estimation algorithm, extensive experiments are conducted. Each case utilizes 600 images for training and 200 images for testing. The sequence length of pressure data is set to 64, with values sampled every 0.5~seconds. To enhance the generalization performance of the algorithm, the corresponding acoustic ranging data for each image is generated following a Gaussian distribution $r_\text{a}\sim N\left(\mu_\text{a},\sigma_\text{a}^2\right)$, as described in Section~\ref{OAFM}.
\begin{figure}[thbp]
\centering
\subfigure[]{
\label{oap_4d}
\includegraphics[width=0.23\textwidth, angle=0]{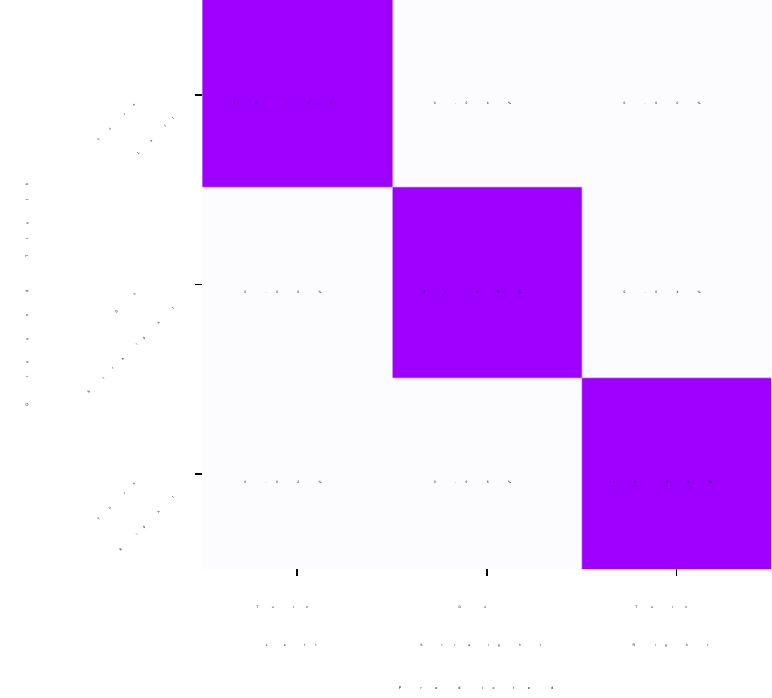}}
\subfigure[]{
\label{oa_4d}
\includegraphics[width=0.23\textwidth]{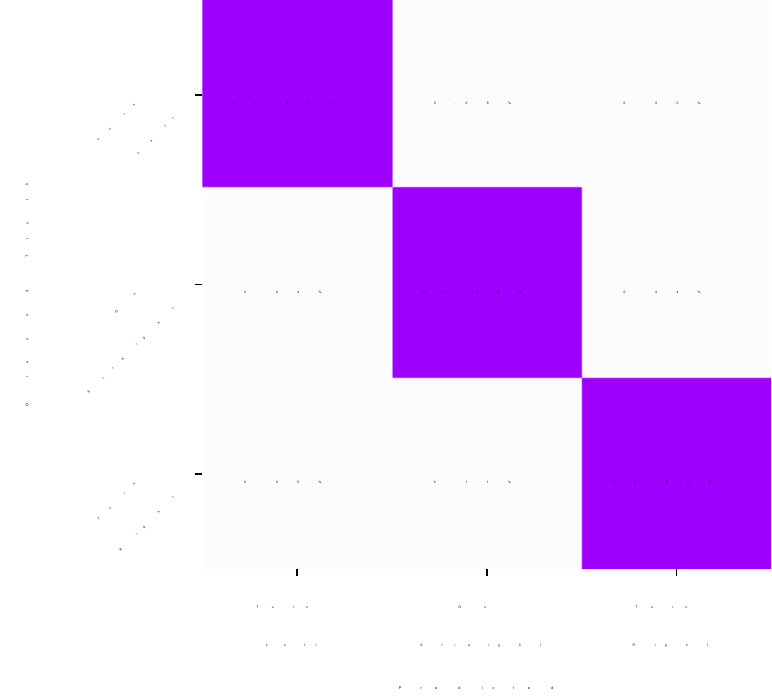}}
\caption{Confusion matrices of estimation result $\hat{d}$ using different modalities. For cases where $p_y\in\{70,90\}$, optical, acoustic and pressure information are used for estimation, the confusion matrix is demonstrated in Fig.~\ref{oap_4d}. For cases where $p_y\in\{170,190,290,310\}$, only optical and acoustic information are used for estimation, the confusion matrix is demonstrated in Fig.~\ref{oa_4d}.}
\label{dir_conf_mat}
\end{figure}
\begin{figure*}[t]
 \centering
 \includegraphics[width=0.9\textwidth]{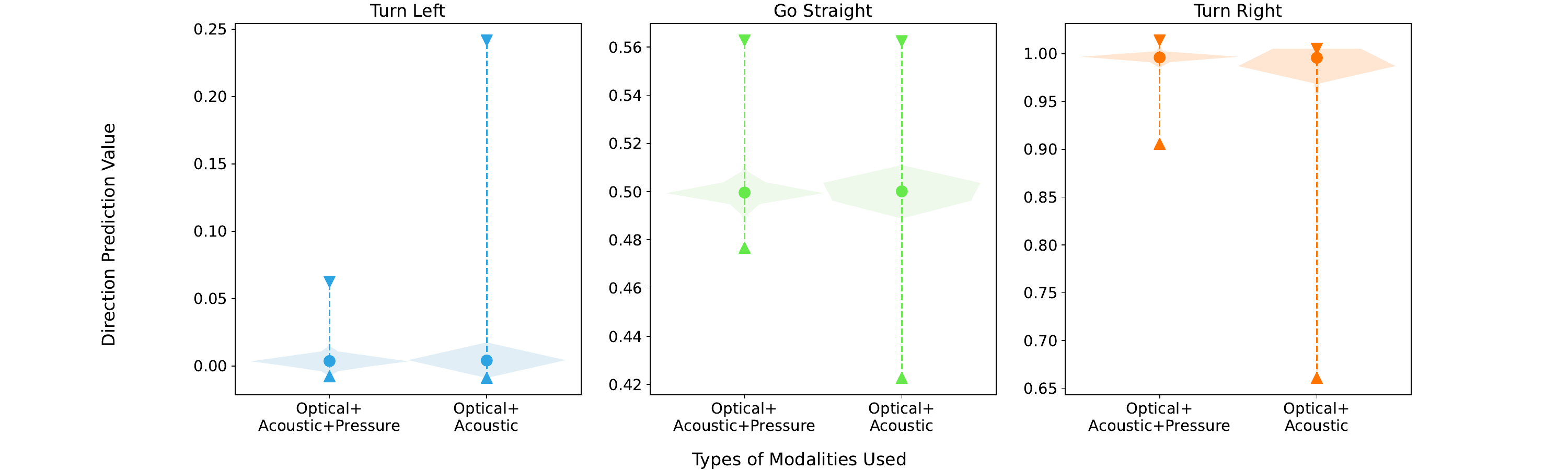}
 \caption{Visualization of the direction state $d$ in a violin plot. The summarized results for `Turn Left', `Move Straight' and `Turn Right' are presented from left to right in three figures with blue, green and orange, respectively. The circle, upper triangle and lower triangle represent the mean value, lower extrema and upper extrema, respectively. Light blue, light green and light orange demonstrate the estimation distributions of the corresponding cases. The horizontal axis represents the modalities used in the estimation process. For cases where $p_y\in\{70,90\}$, optical, acoustic and pressure information are used for estimation and the summary result is demonstrated on the left of each pair. For cases where $p_y\in\{170,190,290,310\}$, only optical and acoustic information are used for estimation and the summary result is demonstrated on the right of each pair. The vertical axis represents the direction state estimates. The ground truth values of `Turn Left', `Move Straight' and `Turn Right' are 0, 0.5 and 1, respectively.}
\label{dir_vis}
\end{figure*}

The regression results for $p_x$ and $p_y$ are visualized in Fig.~\ref{avp_pos_vis} with each circle representing a test instance. Each of the 42 locations is assigned a unique color. Table~\ref{pos_tab} presents the statistical results, listing the Root Mean Square Error (RMSE) and Standard Deviation (SD) for the estimated states $\hat{p_x}$, $\hat{p_y}$, and $\hat{d}$. For $p_y\in\{70,90\}$, all three modalities are used, whereas for $p_y\in\{170,190,290,310\}$, only optical and acoustic sensor measurements are available.

For $d$, thresholds are applied to filter out outliers, and the estimation value $\hat{d}$ is classified as `Turn Left' ($\hat{d}<=0.25$), `Move Straight' ($0.25<\hat{d}<0.75$) and `Turn Right' ($\hat{d}>=0.75$). The confusion matrices, illustrated in Fig.~\ref{dir_conf_mat}, demonstrate that precision and recall values are both over $99\%$ for all cases, whether two or three modalities are used. Incorporating the pressure modality yields an average improvement of $0.03\%$ among all three actions. The mean values, upper and lower extrema are listed in Table \ref{mean_extrema}. 

The Floating Point Operations (FLOPs) and Parameters (Params) of the proposed tri-modal fusion network are 98.17~M and 1.73~M, respectively. Based on the FLOPs and the Params, the proposed architecture is rather lightweight and thus well-suited for deployment on edge devices.

\begin{table*}[t]
\caption{Statistical Results of Proposed Method}
\vspace{-5pt}
\begin{center}
\renewcommand\arraystretch{1.4}
\begin{threeparttable}
\begin{tabular}{cccccccc}
\toprule
\multirow{2}*{Modality} & \multirow{2}*{Condition} & \multicolumn{2}{c}{$\hat{p_x}$} & \multicolumn{2}{c}{$\hat{p_y}$} & \multicolumn{2}{c}{$\hat{d}$} \\ \cmidrule(lr){3-4}\cmidrule(lr){5-6}\cmidrule(lr){7-8} & & RMSE & SD & RMSE & SD & RMSE & SD \\
\midrule
Optical+Acoustic+Pressure & $p_y\in\{70,90\}$ & $0.00464$ & $0.00372$ & $0.00339$ & $0.00336$ & $0.00761$ & $0.00699$\\
Optical+Acoustic & $p_y\in\{170,190,290,310\}$ & $0.00498$ & $0.00466$ & $0.00444$ & $0.00435$ & $0.00859$ & $0.00818$\\
Overall & $p_y\in\{70,90,170,190,290,310\}$ & $0.00487$ & $0.00437$ & $0.00412$ & $0.00405$ & $0.00803$ & $0.00766$\\
\bottomrule
\end{tabular}
\begin{tablenotes}
\footnotesize
\item The RMSE and SD values are calculated using normalized values per Section~\ref{SE}.
\end{tablenotes}
\end{threeparttable}
\end{center}
\label{pos_tab}
\end{table*}

\begin{table*}[t]
\caption{Mean, upper and lower extrema for direction prediction}
\vspace{-5pt}
\begin{center}
\renewcommand\arraystretch{1.4}
\begin{threeparttable}
\begin{tabular}{ccccccc}
\toprule
\multirow{3}*{Action} & \multicolumn{2}{c}{Mean} & \multicolumn{2}{c}{Lower Extrema} & \multicolumn{2}{c}{Upper Extrema} \\
\cmidrule(lr){2-3}\cmidrule(lr){4-5}\cmidrule(lr){6-7} & \multirow{2}*{Optical+Acoustic} & Optical+Acoustic & \multirow{2}*{Optical+Acoustic} & Optical+Acoustic & \multirow{2}*{Optical+Acoustic} & Optical+Acoustic\\
 & & +Pressure & & +Pressure & & +Pressure \\
\midrule
Turn Left & $0.00390$ & $0.00364$ & $-0.00821$ & $-0.00791$ & $0.20792$ & $0.07421$ \\
Move Straight & $0.49967$ & $0.50007$ & $0.32362$ & $0.38731$ & $0.71077$ & $0.57105$\\
Turn Right & $0.99589$ & $0.99591$ & $0.82180$ & $0.89858$ & $1.00519$ & $1.01247$\\
\bottomrule
\end{tabular}
\begin{tablenotes}
\footnotesize
\item The mean, upper and lower extrema are normalized values per Section~\ref{SE}.
\item When predicting with three modalities, all three mean values are closer to ground truth, all three lower extrema are larger and two out of three upper extrema are smaller, indicating a better overall prediction performance.
\end{tablenotes}
\end{threeparttable}
\end{center}
\label{mean_extrema}
\end{table*}

\begin{table*}[ht]
\caption{Statistical Results of Ablation Study and Comparison Study}
\vspace{-5pt}
\begin{center}
\renewcommand\arraystretch{1.4}
\begin{threeparttable}
\begin{tabular}{ccccccccc}
\toprule
\multirow{2}*{Model} & \multirow{2}*{Modality} & \multirow{2}*{Condition} & \multicolumn{2}{c}{$\hat{p_x}$} & \multicolumn{2}{c}{$\hat{p_y}$} & \multicolumn{2}{c}{$\hat{d}$} \\ \cmidrule(lr){4-5}\cmidrule(lr){6-7}\cmidrule(lr){8-9} & & & RMSE & SD & RMSE & SD & RMSE & SD \\
\midrule
Ours & Pressure & $p_y\in\{70,90\}$ & $0.02221$ & $0.02100$ & $0.01846$ & $0.01834$ & $0.16593$ & $0.14242$\\
Ours & Optical & $p_y\in\{70,90\}$ & $0.00641$ & $0.00619$ & $0.00630$ & $0.00555$ & $0.01129$ & $0.01119$\\
Ours & Optical & $p_y\in\{170,190,290,310\}$ & $0.00664$ & $0.00638$ & $0.00679$ & $0.00592$ & $0.01165$ & $0.01136$\\
Ours & Optical Overall & $p_y\in\{70,90,170,190,290,310\}$ & $0.00657$ & $0.00632$ & $0.00665$ & $0.00581$ & $0.01156$ & $0.01131$\\
\midrule
Baseline 1 & Overall & $p_y\in\{70,90,170,190,290,310\}$ & $0.00553$ & $0.00531$ & $0.00570$ & $0.00539$ & $0.01042$ & $0.01031$ \\
Baseline 2 & Overall & $p_y\in\{70,90,170,190,290,310\}$ & $0.00508$ & $0.00479$ & $0.00489$ & $0.00467$ & $0.00884$ & $0.00845$ \\
\bottomrule
\end{tabular}
\begin{tablenotes}
\footnotesize
\item The RMSE and SD values are calculated using normalized values per Section~\ref{SE}.
\end{tablenotes}
\end{threeparttable}
\end{center}
\label{dir_tab}
\end{table*}

\subsection{Ablation Study}
To highlight the advantages of the proposed tri-modal fusion method, this section presents comparison experiments, contrasting leader localization results using tri-modal and single-modal sensory data.

\subsubsection{Single-modal-based State Prediction}
In this section, pure RGB-based and pressure based state estimation results are summarized and analyzed.

\textit{RGB-Based State Estimation:} 
The analysis covers all 126 cases, corresponding to $p_y\in\{70,90,170,190,290,310\}$. The OAFEM backbone structure is employed for model training and testing. The model input consists of RGB images, while the output matches that of the proposed algorithm, comprising three regression states. The training and testing environment, as well as the hyperparameters, remain the same. Statistical results are summarized in Table \ref{dir_tab}.

\textit{Pressure-Based State Estimation:} 
This analysis is limited to cases where pressure data is effective, specifically for $p_y\in\{70,90\}$. The PFEM backbone structure is utilized for the training and testing process. The model input consists solely pressure measurement sequences. The training and testing environment, as well as the hyper-parameters are kept consistent. Statistical results are detailed in Table \ref{dir_tab}.

Using optical data alone, the RMSE and SD values for all three states are approximately 1.5 times larger than their tri-modal counterparts. With pressure data alone, the estimation results show significantly larger errors compared to the tri-modal approach. For $\hat{p_x}$ and $\hat{p_y}$, the RMSE and SD are 4 to 6 times higher, while for $\hat{d}$, these values are 20 times greater.

\subsubsection{Dual-modal-based State Prediction} For cases where $p_y\in\{170,190,290,310\}$, only optical and acoustic information, i.e. RGBA image, is used for state prediction. For all three states, the RMSE and SD values increase by an average of $17.06\%$ and $23.92\%$, respectively, when pressure modality is unavailable, as shown in Table \ref{pos_tab}.

The inclusion of pressure information reduces RMSE and SD by an average of $5.31\%$ and $6.49\%$, respectviely, for all estimated states. Among them, $\hat{p_y}$ achieves the lowest RMSE (0.00412) and SD (0.00405), followed closedly by $\hat{p_x}$. The largest RMSE and SD values are observed for $\hat{d}$, approximately twice those of the other states, yet they remain within an acceptable range, as illustrated in Figs.~\ref{dir_conf_mat} and \ref{dir_vis}.

This comprehensive comparison of state estimation results using single-modal, dual-modal, and tri-modal inputs verifies the effectiveness of the proposed algorithm.

\subsection{Comparison Study}
To show the superiority of our tri-modal fusion architecture, we conducted comparison experiments to compare our framework against other fusion architectures. The two baselines used to compare are specified below. The environment and hyper-parameter used during training and testing stage remain the same.

{\bf Baseline 1} We replaced our hybrid pressure feature extraction module (1D CNN + BiLSTM) with a pure 1D CNN for pressure sequence feature extraction \cite{Xu2022}, \cite{Rodwell2023}.

{\bf Baseline 2} We modified the fusion strategy from our feature-level concatenation to decision-level fusion (late fusion) to align with some multi-sensor fusion paradigms \cite{Vohra_2023_CVPR}, \cite{Ebner2024ICRA}.

The RMSE and SD results for all three estimation states ($p_x, p_y, d$) are summarized in Table~\ref{dir_tab}. Our tri-modal method achieves superior performance, with RMSE reductions of 11.93\% for $p_x$, 27.72\% for $p_y$, and 22.94\% for $d$ compared to Baseline 1, and RMSE reductions of 4.13\% for $p_x$, 15.75\% for $p_y$, and 9.16\% for $d$ compared to Baseline 2. Similarly, the SD values for all states are consistently lower, demonstrating improved robustness. These results validate the effectiveness of our hybrid spatiotemporal pressure feature extraction and feature-level fusion design.

\subsection{Error Analysis}

The localization accuracy is influenced by inherent limitations of individual sensor modalities. Optical sensors, while providing high-resolution imaging, suffer from reduced clarity in turbid water, motion blur from dynamic currents, and a narrow field of view (FOV), which degrades feature extraction at longer distances (e.g., RMSE increases by 7.2\% for optical-only cases at $p_y>=170$cm). Acoustic sensors, though effective for long-range detection, exhibit angular dependency and susceptibility to multi-path interference, leading to higher variance in ranging measurements (e.g., increased SD for $d$ by 23.9\% in dual-modal setups). Pressure sensors, while offering depth and wake flow context, are limited by ambient turbulence and sparse spatial resolution, rendering them ineffective beyond $p_y=90$cm due to indistinguishable noise.

The tri-modal fusion framework mitigates these limitations by synergizing complementary sensor strengths. Optical-acoustic fusion via RGBA heatmaps enhances visual attention in low-visibility scenarios, improving localization accuracy by 17.1 at $p_y>=170$cm compared to optical-only methods. Pressure data integration reduces RMSE by 5.3\% at close ranges. The hybrid CNN-BiLSTM architecture further extracts robust spatiotemporal features from pressure sequences, while the Smooth $L_1$ loss balances outliers, ensuring stability in dynamic conditions.

\subsection{Field Test}

\begin{figure}[ht]
 \centering
 \includegraphics[width=0.45\textwidth]{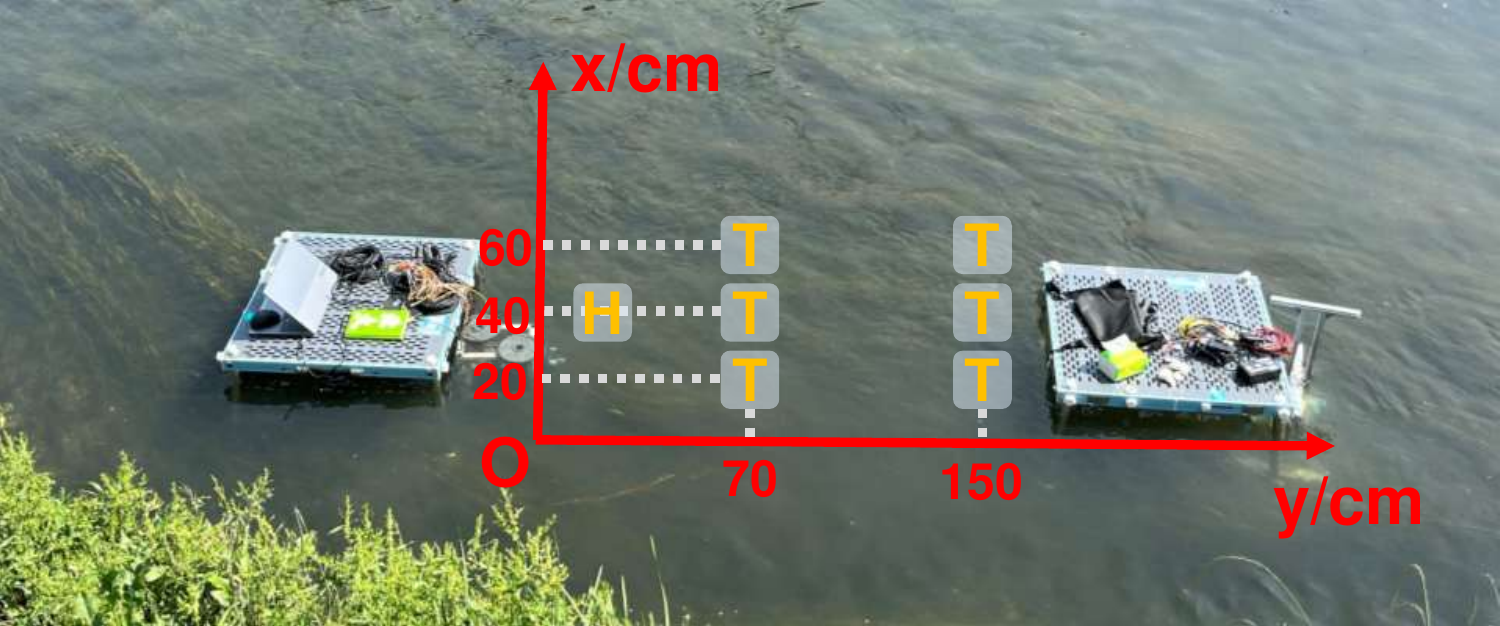}
 \caption{Illustration of the 6 sample locations (top view) in Qinghe River, Beijing. The fixed multi-modal sensing module is denoted as \textbf{H}, while the target module, represented as \textbf{T}, is manually positioned at each location. For each location, three orientations of the target module are tested.}
\label{fieldtest}
\end{figure}

We designed and conducted a field test in Qinghe River, Beijing. The river depth measures 65~cm. To accommodate to the river condition, we make several adjustments to our test platform and test procedure. Instead of 12V DC power supply in lab, we use 12V rechargeable lithium battery packs as power supply for the pair of propellers on the target module. Considering the battery limitation, we sampled at 6 locations with 3 orientation each location. The sampled locations are illustrated in Fig.~\mbox{\ref{fieldtest}}.

\begin{figure}[thpb]
 \centering
 \includegraphics[width=0.4\textwidth]{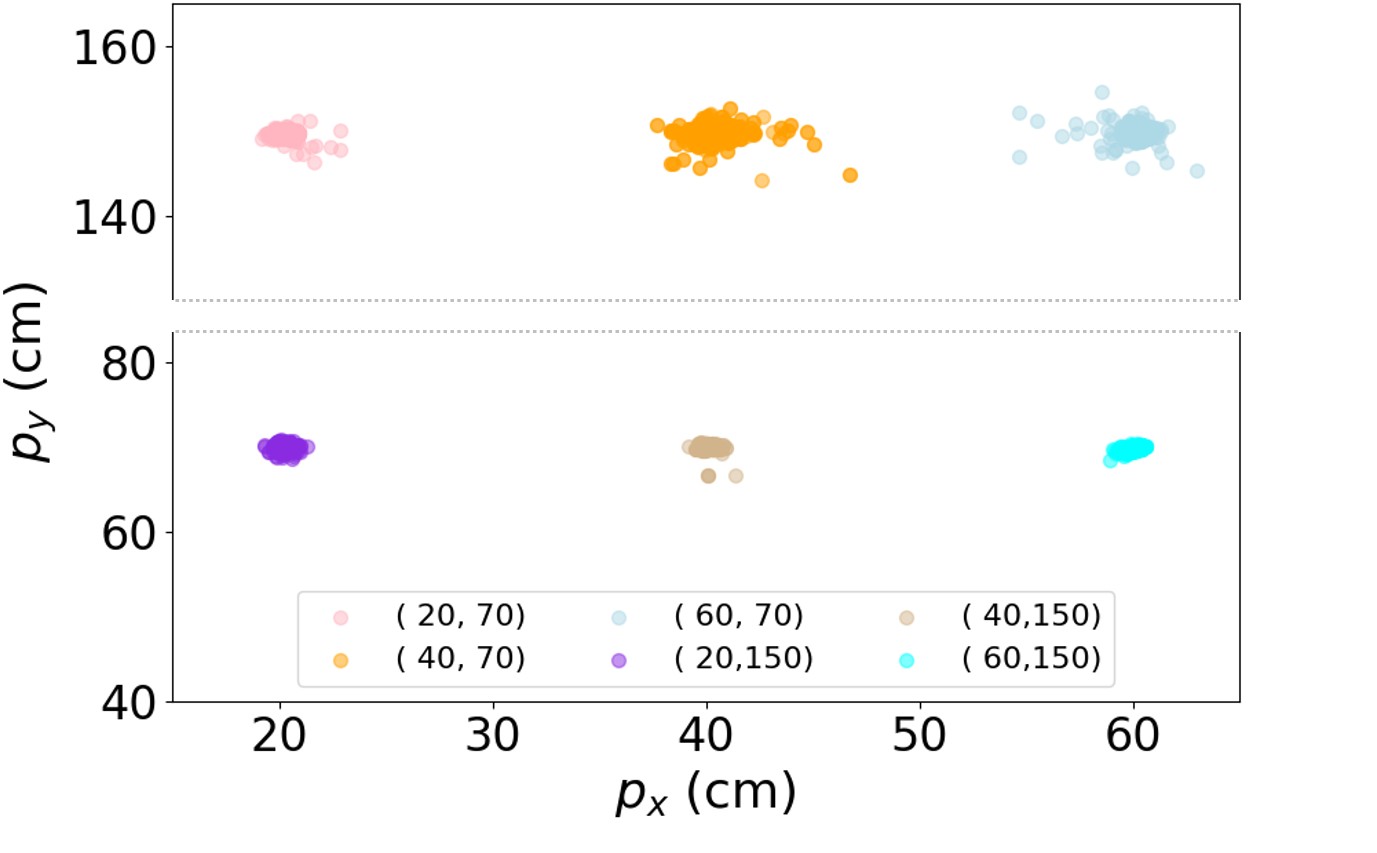}
 \caption{Visualization of position states $(p_x, p_y)$ in a scatter plot. The horizontal and vertical axes represent the $x$-axis and $y$-axis, respectively.}
\label{field-loc-scatter}
\end{figure}

The training and testing environment and process remain the same. The location prediction result is visualized in scatter plot, as shown in Fig.~\mbox{\ref{field-loc-scatter}}. The RMSE and SD values for $\hat{p_x}$ are 0.00735 and 0.00672, respectively while those for $\hat{p_y}$ are 0.00750 and 0.00721, respectively. The RMSE and SD values for $\hat{d}$ are 0.01142 and 0.01096, respectively. Consequently, the proposed method is capable of handling the localization task under various underwater environments. Due to the complex interference (i.e. turbidity and currents) in the river environment, the RMSE and SD values of the $\hat{p_x}, \hat{p_y}$ and $\hat{d}$ are larger than the RMSE and SD values of the $\hat{p_x}, \hat{p_y}$ and $\hat{d}$ in the laboratory-based testing platform but within the acceptable range, proving the robustness of the proposed method.

\section{Conclusion}
This paper presented a tri-modal fusion approach to address the critical challenge of localizing a leader vehicle within multiple-vehicle systems operating in demanding underwater environments. By integrating optical, acoustic, and pressure sensor measurements within a deep neural network framework, the proposed method achieved significant improvements in state estimation accuracy and robustness. The fusion process was systematically structured into two stages. At the data level, acoustic and optical information are combined by constructing an RGB-Attention image. This image integrated a heatmap, representing the Gaussian distribution of acoustic sensor data, with the RGB camera image along the channel dimensions. At the feature level,  optical-acoustic and pressure feature were extracted using convolutional and LSTM blocks and subsequently fused to generate leader state estimates.
A comprehensive testing platform was developed to evaluate the proposed approach. Extensive experimental results, including comparison studies and field tests, validated its effectiveness.

For future research, we plan to enhance the sensor fusion algorithm, integrate new sensor modalities and test the method through extended applications. The algorithmic enhancement for sensor fusion incorporates dynamic weight adaptation, temporal-spatial modeling and online learning. We also attempt to solve challenges involving real-time performance optimization and utilization of energy efficiency. Meanwhile, we will deploy our algorithm on underwater vehicles and evaluate the algorithm under dynamic underwater conditions to verify the effectiveness of our method.

\bibliography{ref}

\vfill

\end{document}